\documentclass[twoside,11pt]{article} 
\usepackage{jmlr2e}

\editor{}
\firstpageno{1}
\author{\name Ashish Khetan \email khetan2@illinois.edu\\ 
\name Sewoong Oh \email swoh@illinois.edu\\ 
       \addr Department of Industrial and Enterprise Systems Engineering\\
       University of Illinois at Urbana-Champaign\\
       Urbana, IL 61801, USA
       }

\usepackage{amssymb,amsfonts,amsmath,amsthm,amscd,dsfont,mathrsfs}
\usepackage{graphicx,float,psfrag,epsfig}
\usepackage{wrapfig}
\usepackage{pgf,pgfarrows,pgfnodes}

\DeclareMathAlphabet{\mathpzc}{OT1}{pzc}{m}{it}

\usepackage{epstopdf}
\usepackage{caption}
\usepackage{subcaption}
\usepackage{array}
\newcolumntype{C}[1]{>{\centering\let\newline\\\arraybackslash\hspace{0pt}}m{#1}}

\usepackage{epstopdf}

\usepackage{array}

\newtheorem{propo}{Proposition}[section]
\newtheorem{lemma}[propo]{Lemma}

\newtheorem{theorem}[propo]{Theorem}

\newtheorem{remark}[propo]{Remark}

\def\U{\mathcal{U}}

\def\C{\mathcal{C}}
\def\G{\mathcal{G}}
\def\Z{\mathbb{Z}}
\def\cG{\mathcal{G}}

\def\tl{{\tilde{\ell}}}
\def\tm{{\tilde{m}}}
\def\tr{{\tilde{r}}}
\def\tell{{\tilde{\ell}}}
\def\L{\mathcal{L}}

\def\cG{\mathcal{G}}
\def\cL{\mathcal{L}}
\def\Lrb{\mathcal{L_{\rm RB}}}
\def\I{\mathbb{I}}

\def\i{i^\prime}

\def\RB{{\rm RB}} 

\def\cS{{\mathcal{S}}}
\def\cP{{\mathcal{P}}}
\def\Tr{{ \rm Tr}}

\def\E{{\mathbb{E}}}
\def\P{{\mathbb{P}}}
\def\H{{\mathcal{H}}}
\def\reals{{\mathbb{R}}}
\def\prob{{\mathbb{P}}}
\def\<{\langle}
\def\>{\rangle}

\def\htheta{\widehat{\theta}}

\def\ltheta{{\widetilde{\theta}}}

\def\lalpha{{\widetilde{\alpha}}}

\newcommand{\norm}[1]{\|#1\|}
\newcommand{\vect}[1]{\boldsymbol{#1}} 
\newcommand{\ceil}[1]{\left \lceil{#1} \right \rceil}
\newcommand{\floor}[1]{\left \lfloor{#1} \right \rfloor}

\begin{document}

\title{Computational and Statistical Tradeoffs in Learning to Rank}



\maketitle


\begin{abstract}
For massive and heterogeneous modern  datasets, it is of fundamental interest to 
provide guarantees on the accuracy of estimation when computational resources are limited. 
In the application of 
learning to rank, we provide a hierarchy of  rank-breaking mechanisms 
ordered by the complexity in thus generated sketch of the data.  
This allows the number of data points collected to be gracefully traded off 
against computational resources available, while guaranteeing the desired level of accuracy. 
Theoretical guarantees on the proposed generalized rank-breaking implicitly provide such trade-offs, 
which can be explicitly characterized under certain 
canonical scenarios  on the structure of the data. 
\end{abstract}

\section{Introduction}
\label{sec:introduction}

In classical statistical inference, we are typically interested in characterizing  how more  
data points improve the accuracy,  with little 
restrictions or considerations on  computational aspects of solving the inference problem. 
However,  with massive growths of the amount of data available and also the complexity and heterogeneity of the 
 collected data, computational resources, such as time and memory, are major bottlenecks in many modern applications.
As a solution, recent advances in learning theory   
introduce  hierarchies of algorithmic solutions, 
ordered by the respective computational complexity, for several fundamental machine learning applications in \citep{BB08,SS08,CJ13,ABD12,LOK15}. 
Guided by 
sharp analyses on the sample complexity, 
these approaches provide theoretically sound guidelines 
that  allow the analyst the flexibility to fall back to simpler algorithms 
to enjoy the full merit of the improved run-time. 

%

Inspired by  these advances, we study the time-data tradeoff in learning to rank. 
In many applications such as election, policy making, polling, and recommendation systems, 
we want to 
aggregate individual preferences to produce a 
global ranking that best represents the collective  social preference.
Learning to rank is a rank aggregation approach, which assumes that the data comes from a parametric family of choice models, 
and learns the parameters that determine the global ranking. 
Traditionally, each revealed preference is assumed to have  one of the following three structures.  
{\em Pairwise comparison}, where one item is preferred over another, is common in sports and chess matches. 
{\em Best-out-of-$\kappa$ comparison}, where one is chosen among a set of $\kappa$ alternatives, is common in 
historical purchase data. 
{\em $\kappa$-way comparison}, 
where we observe a linear ordering of a set of $\kappa$ candidates, is used in some elections and  surveys. 
We will refer to such structures as {\em traditional} in comparisons to 
modern datasets with non-traditional structures whose behavior change drastically. 
 For such traditional preferences, efficient schemes for learning to rank have been proposed, 
such as \cite{Ford57,Hun04,HOX14,CS15}, which we explain in detail in Section \ref{sec:related}.
However, modern datasets are unstructured and heterogeneous. 
As \cite{KO16} show, this can lead to significant increase in the computational complexity, requiring exponential run-time in the 
size of the problem in the worst case. 

To alleviate this computational challenge, 
we propose a hierarchy of estimators which we call {\em generalized rank-breaking}, 
ordered in increasing computational complexity and 
achieving  increasing accuracy. 
The key idea is to break down the heterogeneous revealed preferences into simpler pieces of ordinal relations, 
and apply an estimator tailored for those simple structures treating each piece as independent. 
Several aspects of rank-breaking makes this problem interesting and challenging. 
A priori, it is not clear which choices of the simple ordinal relations are rich enough to be statistically efficient 
and yet lead to tractable estimators. 
Even if we identify which ordinal relations to extract, 
the ignored correlations among those pieces can lead to an inconsistent estimate, 
unless we choose carefully  which  pieces to include and which to omit in the estimation. 
We further want sharp analysis on the sample complexity, 
which reveals how computational and statistical efficiencies trade off. 
We would like to address all these challenges in providing generalized rank-breaking methods.  

\bigskip\noindent
{\bf Problem formulation.} 
We study the problem of aggregating ordinal data based on users' preferences that are expressed in the form of {\emph{partially ordered sets (poset)}}. 
  A poset is a collection of ordinal relations among items. 
  For example, consider 
  a poset $\{ (i_6\prec\{i_5,i_4\}),(i_5\prec i_3),(\{i_3,i_4\}\prec\{i_1,i_2\} ) \}$ 
  over items $\{i_1,\ldots,i_6\}$, where $(i_6\prec \{i_5,i_4\})$ indicates that item $i_5$ and $i_4$ are both preferred over item $i_6$. Such a relation is extracted from, for example, the user giving a 2-star rating to  $i_5$ and $i_4$ and a 1-star to $i_6$. 
 Assuming that the revealed preference is consistent, a poset can be represented as a directed acyclic graph (DAG) $\G_j$ as below.

 \begin{figure}[h]
 	\begin{center}
	\includegraphics[width=.6\textwidth]{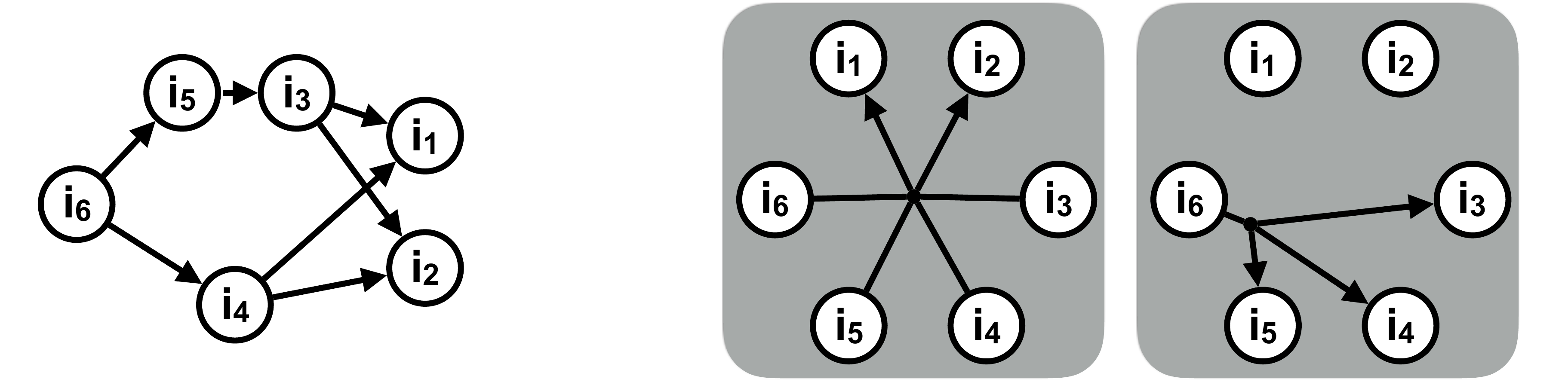}
	\put(-240,69){$\cG_j$}
	\put(-122,71){$e_1$}
	\put(-52,71){$e_2$}
	\end{center}
	\caption{An example of $\cG_j$ for  user $j$'s consistent poset, and  
	two rank-breaking hyper edges extracted from it: $e_1=(\{i_6,i_5,i_4,i_3\} \prec \{i_2,i_1\})$ and $e_2=(\{i_6\} \prec \{i_5,i_4,i_3\})$. }
	\label{fig:example}
 \end{figure}

We assume that each user $j$ is presented with a subset of items $S_j$, and independently provides her ordinal preference 
in the form of a poset, where the ordering is drawn from the 
Plackett-Luce (PL) model. 
The PL model is a popular choice model from operations research and psychology,  
used to model how people make choices under uncertainty. 
It is a special case of {\em random utility models}, where 
each item $i$ is parametrized by a  latent true utility $\theta_i\in\reals$.  
When offered with $S_j$, 
the user  samples the perceived utility $U_i$ for each item independently 
according to 
$U_i =\theta_i+Z_i$, where $Z_i$'s are i.i.d. noise. 
In particular, the PL model assumes $Z_i$'s follow the standard Gumbel distribution. 
The observed poset is a partial observation of the ordering according to this perceived  utilities. 
We discuss possible extensions to general class of random utility models in Section \ref{sec:related}.

The particular choice of the Gumbel distribution has several merits, largely stemming from the fact that the Gumbel distribution has a log-concave pdf and is inherently memoryless. 
In our analyses, we use the log-concavity to show that our proposed algorithm  is a concave maximization (Remark \ref{lem:concave}) and the memoryless property forms the basis of our rank-breaking idea. Precisely, the PL model is statistically equivalent to the following procedure. 
Consider a ranking as a mapping from a position in the rank to an item, i.e. 
$\sigma_j : [|S_j|] \rightarrow S_j$.   
It can be shown that the PL model  is generated by 
first  independently assigning each item $i \in S_j$ an unobserved value $Y_i$, exponentially distributed with mean $e^{-\theta_i}$, and the resulting ranking $\sigma_j$ is inversely ordered in $Y_i$'s so that $Y_{\sigma_j(1)} \leq Y_{\sigma_j(2)}\leq \cdots \leq Y_{\sigma_j(|S_j|)}$. 

This inherits the memoryless property of exponential variables, such that 
$\prob(Y_1 < Y_2 < Y_3)=\prob(Y_1<\{Y_2,Y_3\}) \prob(Y_2<Y_3)$, leading to a simple  interpretation of the PL model as sequential choices: 
$$\prob(i_3\prec i_2\prec i_1)\;=\;\prob(\{i_3,i_2\} \prec i_1)\prob(i_3\prec i_2) 
	\;=\; \frac{e^{\theta_{i_1}}}{e^{\theta_{i_1}}+e^{\theta_{i_2}}+e^{\theta_{i_3}} } \times \frac{e^{\theta_{i_2} }}{e^{\theta_{i_2}}+e^{\theta_{i_3}} } \;. $$
In general, we have 
 $\P[\sigma_j] = \prod_{i=1}^{|S_j|-1}(e^{\theta^*_{\sigma_j(i)}})/(\sum_{i'=i}^{|S_j|}e^{\theta^*_{\sigma_j(i')}})$. We assume that the true utility $\theta^* \in \Omega_b$ where \begin{eqnarray}
 	\Omega_b \;=\; \Big\{\, \theta \in \reals^d \,\big|\, \sum_{i \in [d]}\theta_i = 0, |\theta_i| \leq b \text{ for all } i \in [d] \, \Big\} \; .
\end{eqnarray}
	 Notice that centering of $\theta$ ensures its uniqueness as PL model is invariant under shifting of $\theta$. The bound $b$ on $\theta_i$ is written explicitly to capture the dependence in our main results.     

We denote a set of  $n$ users by $[n]=\{1,\ldots,n\}$ and 
the set of $d$ items by $[d]$. 
Let $\G_j$ denote the DAG representation of the poset provided by the user $j$ over $S_j \subseteq [d]$ according to the PL model with weights $\theta^*$.  
The maximum likelihood estimate (MLE) maximizes 
the sum of all possible rankings that are consistent with the observed $\G_j$ for each $j$: 
\begin{eqnarray}
	\label{eq:est1}
	\widehat{\theta} &\in  &\arg \max_{\theta \in \Omega_b} \bigg\{ \sum_{j=1}^n \log 	\bigg( \sum_{\sigma \in \G_j}\P_{\theta}[\sigma] \bigg)\bigg\}\,,
\end{eqnarray} 
where we slightly abuse the notation $\cG_j$ to denote the set of all rankings $\sigma$ that 
are consistent with the observation.
 When $\G_j$ has a {\em traditional} structure as explained earlier in this section, then the optimization is a simple multinomial logit regression, that can be solved efficiently with off-the-shelf convex optimization tools. \cite{HOX14} provides full analysis of the statistical complexity of this MLE under traditional structures. 
For general posets, it can be shown that the above optimization is a concave maximization, using similar techniques as 
Remark \ref{lem:concave}.
However,  
the summation over rankings in $\G_j$ can involve number of terms super exponential in the size $|S_j|$, in the worst case. 
This renders MLE  intractable and impractical. 

\bigskip\noindent
{\bf Pairwise rank-breaking.}
A common remedy to this computational blow-up is to use rank-breaking. 
Rank-breaking traditionally refers to 
{\em pairwise rank-breaking}, where a bag of all the pairwise comparisons is  extracted 
from observations $\{\cG_j\}_{j\in[n]}$ 
and is applied to estimators that are tailored for pairwise comparisons, 
treating each paired outcome as independent. 
This is one of the motivations 
behind the  algorithmic advances in the popular topic of learning from pairwise comparisons 
in \citep{Ford57,Hun04,NOS14,SBB15,MG15}.

It is computationally efficient to apply maximum likelihood estimator assuming independent pairwise comparisons, which 
takes $O(d^2)$ operations to evaluate. 
However, this computational gain comes at the cost of statistical efficiency. 
\cite{APX14a} showed that if we include all paired comparisons, 
then the resulting estimate can be statistically  inconsistent due to the ignored correlations among the paired orderings, even with infinite samples.   In the example from Figure \ref{fig:example}, there are 12 paired relations implied by the DAG: $(i_6\prec i_5),(i_6\prec i_4),(i_6\prec i_3),\ldots,(i_3\prec i_1),(i_4\prec i_1)$.
In order to get a consistent estimate, 
\cite{APX14a} provide a rule for choosing which pairs to include, 
and \cite{KO16} provide an estimator that optimizes how to weigh each of those chosen pairs to get the best finite sample complexity bound. 
However, 
such a consistent pairwise rank-breaking results in throwing away 
many of the ordered relations, resulting in significant loss in accuracy. 
For example, Including a paired relation from $\G_j$ in the example results in a biased estimator.
None of the pairwise orderings can be used from $\G_j$, without making the estimator inconsistent as shown in \cite{ACPX13}. 
Whether we include all paired comparisons 
or only a subset of consistent ones, there is a significant loss in accuracy as illustrated in Figure \ref{fig:fig1}. 
For the precise condition for consistent rank-breaking we refer to \citep{ACPX13,APX14a,KO16}.

The state-of-the-art approaches operate on  either one of the two extreme points on the computational and statistical trade-off. 
The MLE in \eqref{eq:est1} requires $O(\sum_{j\in[n]} |S_j|!)$ summations to just evaluate the objective function, in the worst case. On the other hand, 
the pairwise rank-breaking requires only  $O(d^2)$ summations, but suffers from significant loss in the sample complexity. 
Ideally, 
we would like to give the analyst the flexibility to choose 
a target computational complexity she is willing to tolerate, 
and provide an algorithm that achieves the optimal trade-off 
at the chosen operating point. 

\bigskip\noindent
{\bf Contribution.}
We introduce a novel {\em generalized rank-breaking} that bridges the gap between MLE and pairwise rank-breaking. Our approach allows the user 
the freedom to choose the level of computational resources to be used, and provides an estimator tailored for the desired complexity. 
We prove that the proposed estimator is tractable and consistent, 
and provide an upper bound on the error rate in the finite sample regime. 
The analysis explicitly characterizes the dependence on the topology of the data. 
This in turn provides a guideline for designing surveys and experiments in practice, 
in order to maximize  the sample efficiency. 
We provide numerical experiments confirming the theoretical guarantees. 

\subsection{Related work}
\label{sec:related} 

In classical statistics, one is interested in the tradeoff between 
the sample size and  the accuracy, with little considerations to the computational complexity or time. 
As more computations are typically required with increasing availability of data, the computational resources are often the bottleneck. 
Recently, a novel idea known as ``algorithmic weakening'' has been investigated to overcome such a bottleneck, in which a hierarchy of algorithms is proposed 
to allow for faster algorithms at the expense of decreased accuracy. 
When guided by sound theoretical analyses, 
this idea allows the statistician to achieve the same level of accuracy and {\em save} time when more data is available. 
This is radically different from classical setting where processing more data typically requires more computational time. 

Depending on the application, several algorithmic weakenings have been studied. 
In the application of supervised learning, 
\cite{BB08} proposed the idea that weaker approximate optimization algorithms are sufficient for learning when more data is available. 
Various gradient based algorithms are analyzed that show the time-accuracy-sample tradeoff. 
In a similar context, 
\cite{SS08} analyze a particular implementation of support vector machine 
and show that the target accuracy can be achieved faster when more data is available, by running the iterative algorithm for shorter amount of time.  
In the application of de-noising, 
\cite{CJ13} provide a hierarchy of convex relaxations where constraints are defined by convex geometry with 
increasing complexity. 
For unsupervised learning, 
\cite{LOK15} introduce a hierarchy of data representations 
that provide more representative elements when more data is available at no additional computation. 
Standard clustering algorithms can be applied to thus generated summary of the data, requiring less computational complexity.

\begin{figure}[h]
 \begin{center}
	\includegraphics[width=.5\textwidth]{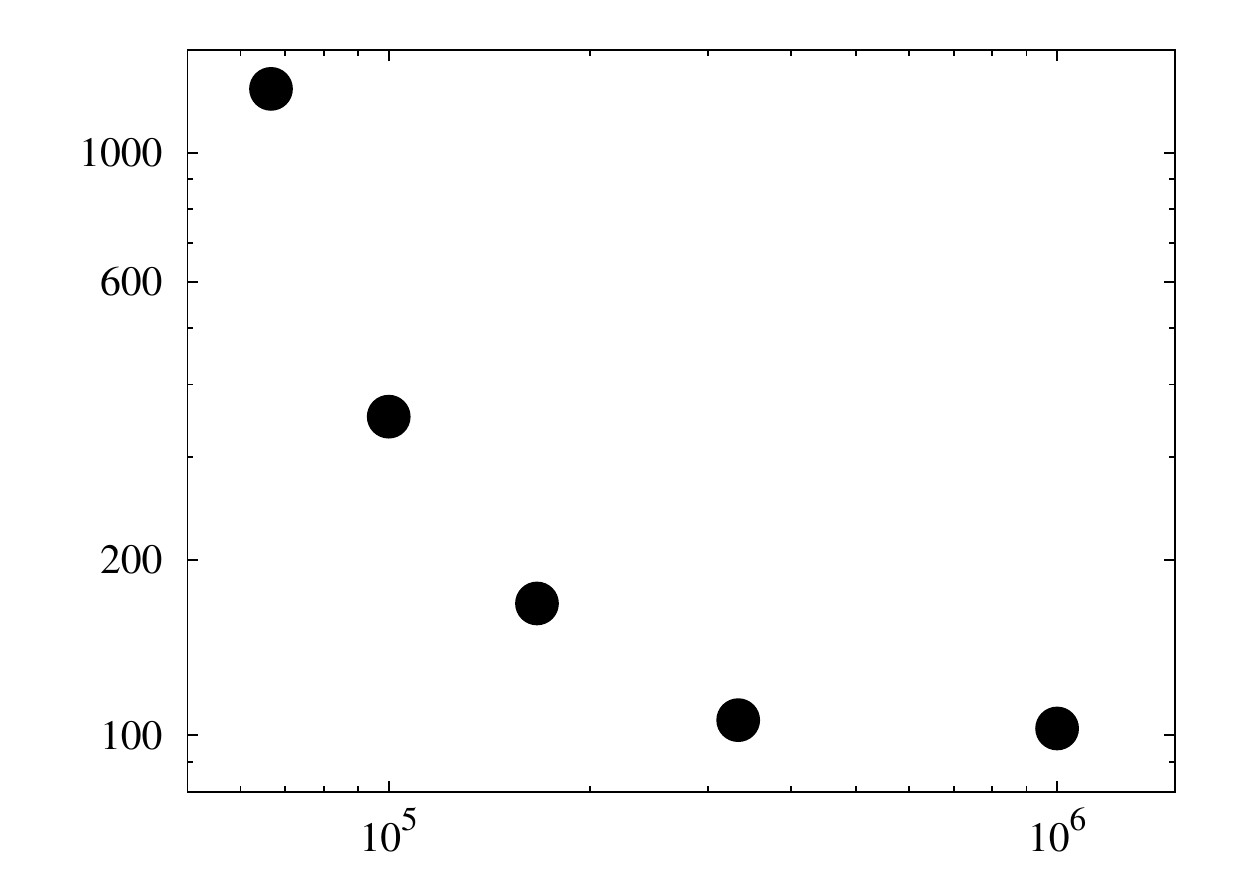}
	\put(-265,107){Time (s)}
	\put(-135,-7){sample size \small{$n$}}
	\put(-32,25){\small{$M=1$}}
	\put(-89,26){\small{$M=2$}}
	\put(-126,48){\small{$M=3$}}
	\put(-157,83){\small{$M=4$}}
	\put(-177,144){\small$M=5$}
\end{center}
\caption{ Depending on how much computational resources are available, the various choices of $M$ achieve 
different operating points on the time-data trade-off to achieve some fixed target accuracy $\varepsilon>0$. 
If more samples are available, one can resort to faster methods with smaller $M$ while achieving the same level of accuracy. }
\label{fig:fig0}
\end{figure}
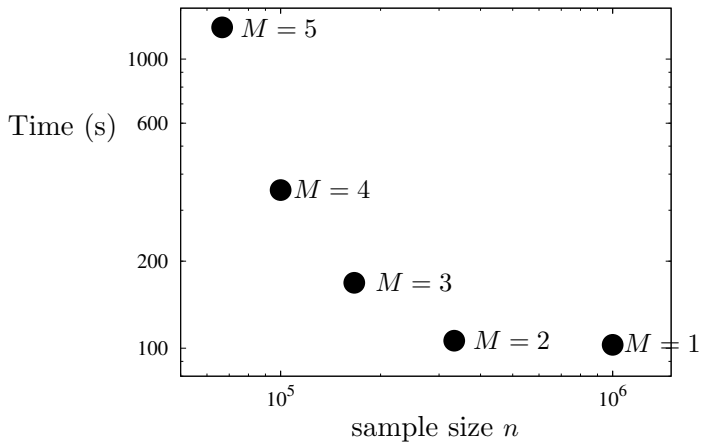

In the application of learning to rank, we follow the principle of algorithmic weakening and propose a novel rank-breaking 
to allow the practitioner to navigate gracefully the time-sample trade off as shown in the figure below. 
We propose a hierarchy of estimators indexed by $M\in\Z^+$ indicating how complex the estimator is (defined formally in Section \ref{sec:grb}). 
Figure \ref{fig:fig0} shows the result of a experiment on synthetic datasets on 
how much time (in seconds) and how many samples are required to achieve a target accuracy. 
If we are given more samples, then it is possible to achieve the target accuracy, which in this example is MSE$\leq 0.3d^2\times \, 10^{-6} $, 
with fewer operations by using a simpler estimator with smaller $M$.
The details of the experiment is explained in Figure \ref{fig:fig1}.

Learning to rank from the PL model has been studied extensively under the {\em traditional} scenario 
dating back to \cite{Zer29} who first introduced the PL model for pairwise comparisons. 
Various approaches for estimating the PL weights from traditional samples have been proposed. 
The problem can be formulated as a convex optimization that can be solved efficiently using the off-the-shelf solvers. 
However, tailored algorithms for finding the optimal solution have been proposed in 
\cite{Ford57} and \cite{Hun04}, which iteratively finds the fixed point of the KKT condition.  
\cite{NOS14} introduce Rank Centrality, a novel spectral ranking algorithm which formulates a random walk 
from the given data, 
and show that the stationary distribution provides accurate estimates of the PL weights. 
\cite{MG15} provide a connection between those previous approaches, and give a unified random walk approach that finds the 
fixed point of the KKT conditions. 

On the theoretical side, 
when samples consist of pairwise comparisons,  \cite{SY99} first established consistency and asymptotic normality of the maximum likelihood estimate  when 
all teams play against each other. 
For a broader class of scenarios where we allow for  sparse observations, where the number of total comparisons grow linearly in the number of teams, 
\cite{NOS14} show that Rank Centrality achieves optimal sample complexity by comparing it to a lower bound on the minimax rate. 
For a more general class of traditional observations, including pairwise comparisons, 
\cite{HOX14} provide similar optimal guarantee for the maximum likelihood estimator. 
\cite{CS15} introduced Spectral MLE that applies Rank Centrality followed by MLE, and showed that 
the resulting estimate is optimal in $L_\infty$ error as well as the previously analyzed $L_2$ error. 
\cite{SBB15} study a new measure of the error induced by the Laplacian of the comparisons graph 
and prove a sharper upper and lower bounds that match up to a constant factor. 

However, in modern applications, 
the computational complexity of the existing approaches blow-up due to the heterogeneity of modern datasets.
Although, 
statistical and computational tradeoffs have been investigated 
under other popular choice models such as the 
Mallows models by \cite{BBN14} or stochastically transitive models by \cite{SBG15}, 
the algorithmic solutions do not apply to random utility models and the analysis techniques do not extend. 
We provide a novel rank-breaking algorithms and provide finite sample complexity analysis under the PL model. 
 This approach readily generalizes to some RUMs such as 
the flipped Gumbel distribution. 
However, it is also known from 
 \cite{APX14a}, 
that for general RUMs there is no consistent rank-breaking, and the proposed approach does not generalize.



\section{Generalized rank-breaking}
\label{sec:grb}

Given $\G_j$'s representing the users' preferences, 
{\em generalized rank-breaking} extracts a set of 
ordered relations  
and applies an estimator treating each ordered relation as independent. 
Concretely, for each $\cG_j$, we first 
 extract a maximal ordered partition $\cP_j$ of $S_j$ 
that is consistent with $\cG_j$. 
An ordered partition is a partition  with a linear ordering among  the subsets, e.g. $\cP_j = ( \{i_6\}\prec \{i_5,i_4,i_3\} \prec \{i_2,i_1\})$ for $\G_j$ from 
Figure \ref{fig:example}.
This is  maximal, 
since we cannot further partition any of the subsets 
without creating artificial ordered relations that are not present in the original $\G_j$.

The extracted ordered partition is represented by 
a directed hypergraph $G_j(S_j,E_j)$, which we call a {\em rank-breaking graph}.  
Each edge $e=(B(e),T(e))\in E_j$ is a directed hyper edge  from a subset of nodes $B(e)\subseteq S_j$ to another subset $T(e)\subseteq S_j$. 
The number of edges in $E_j$ is $|\cP_j|-1$ where $|\cP_j|$ is 
the number of subsets in the partition.
For each subset in $\cP_j$ except for the least preferred subset, there is a corresponding edge whose {\em top-set} $T(e)$ is the subset, and the 
  {\em bottom-set} $B(e)$ is the set of all items less preferred than $T(e)$. 
For the example in Figure \ref{fig:example}, we have $E_j=\{e_1,e_2\}$ where  
$e_1=(B(e_1),T(e_1))=(\{i_6,i_5,i_4,i_3\},\{i_2,i_1\})$ and 
$e_2=(B(e_2),T(e_2)=(\{i_6\},\{i_5,i_4,i_3\})$ extracted from $\G_j$.
Denote the probability that $T(e)$ is preferred over $B(e)$ when $T(e)\cup B(e)$ is offered as 
\begin{eqnarray}
	\prob_{\theta}( e) \;=\; \prob_\theta \big(B(e)\prec T(e) \big) \;=\; \sum_{\sigma \in \Lambda_{T(e)}} \frac{\exp\left(\sum_{c=1}^{|T(e)|}\theta_{\sigma(c)}\right)}{\prod_{u=1}^{|T(e)|} \left( \sum_{c' = u}^{|T(e)|} \exp\left(\theta_{\sigma(c')} \right) + \sum_{i\in B(e)} \exp\left( \theta_i \right) \right)}
	\label{eq:pair}
\end{eqnarray} 
which follows from the definition of the PL model, 
where $\Lambda_{T(e)}$ is the set of all rankings over $T(e)$. 
The computational complexity of evaluating this probability 
is determined  by the size of the {\em top-set} $|T(e)|$, as it involves  $(|T(e)|!)$ summations. 

We let the analyst choose the order $M\in\Z^+$ depending on how much computational resource is available, and only include those edges with $|T(e)|\leq M$ in the following step. 
We apply the MLE for comparisons over paired subsets, 
assuming all rank-breaking graphs are independently drawn.   
Precisely, we propose {\em order-$M$ rank-breaking estimate}, which is  the solution that maximizes the log-likelihood under the independence assumption:  
\begin{eqnarray}
	\label{eq:estimate}
	\widehat{\theta} &\in & \arg\max_{\theta \in \Omega_b} \L_{\RB}(\theta)\,\;,\text{ where }\;\;\;\;\;\;
	\L_{\RB}(\theta) \;=\;  \sum_{j \in [n]}  \sum_{e \in E_j: |T(e)|\leq M} \log \prob_{\theta}( e ) \;. 
	\end{eqnarray}
In a special case when $M=1$, this  can be transformed into  
 the traditional pairwise rank-breaking, where 
$(i)$ this  is a concave maximization; 
$(ii)$ the estimate is (asymptotically) unbiased and consistent as shown in \cite{ACPX13,APX14a}; and 
$(iii)$  and the finite sample complexity have been analyzed in \cite{KO16}.  
Although, this order-$1$ rank-breaking provides a significant gain in computational efficiency, 
the information contained in higher-order edges are unused, resulting in a significant loss in accuracy. 

We provide the analyst the freedom to choose the computational complexity he/she is willing to tolerate.   
However, for general $M$, 
it has not been known  if the optimization in \eqref{eq:estimate} is tractable and/or 
if the solution is consistent. 
Since 
$\prob_{\theta}(B(e) \prec T(e) )$ as explicitly written in \eqref{eq:pair} 
is a sum of log-concave functions, it is not clear if the sum is also log-concave. 
Due to the ignored dependency in the formulation \eqref{eq:estimate}, it is not clear if the resulting estimate is consistent. 
We first establish that it is a concave maximization in Remark \ref{lem:concave},  
then prove consistency in Remark \ref{lem:consistent}, 
and provide a sharp analysis of the performance in the finite sample regime,  characterizing the trade-off between 
computation and sample size in Section \ref{sec:main}.  
We use the Random Utility Model (RUM)   interpretation of the PL model to prove concavity. We refer to Section \ref{sec:concave}   for a proof. 
\begin{remark} 
	\label{lem:concave}
	$\cL_{\rm RB}(\theta)$ is concave in $\theta\in\reals^d$. 
\end{remark}  

In order to discuss consistency of the proposed approach, 
we need to specify how we sample the set of items to be offered $S_j$ and also which partial ordering over $S_j$ is to be observed. 
Here,  we consider a 
 simple but  canonical scenario for sampling ordered relations, and show the proposed method is consistent for all non-degenerate cases. 
However,  we study a more general sampling scenario, when we analyze the order-$M$ estimator in the finite sample regime in Section \ref{sec:main}. 

Following is the canonical sampling scenario. There is a set of $\tl$ integers $(\tm_1,\ldots,\tm_\tl)$ whose sum is strictly less than $d$. A new arriving user is presented with all $d$ items and is asked 
to provide 
her top $\tm_1$ items as an unordered set, 
and then the next $\tm_2$ items, and so on. 
This is sampling from the PL model and observing 
an ordered partition with $(\tl+1)$ subsets of sizes $\tm_a$'s, and the last subset includes all remaining items. We apply the generalized rank-breaking to get rank-breaking graphs $\{G_j\}$ with $\tl$ edges each, and  order-$M$ estimate is computed.  
We show that this  is consistent, i.e. asymptotically unbiased  in the limit of the number of users $n$. 
A proof is provided in Section \ref{sec:consistent}. 
\begin{remark} 
	\label{lem:consistent}
	Under the {\rm{PL}} model and the above sampling scenario, the order-$M$ rank-breaking estimate 
	$\htheta$ in \eqref{eq:estimate} is consistent for all choices of $M\geq \min_{a \in \tl} \tm_a $. 
\end{remark}

\begin{figure}[h]
 \begin{center}
	\includegraphics[width=.48\textwidth]{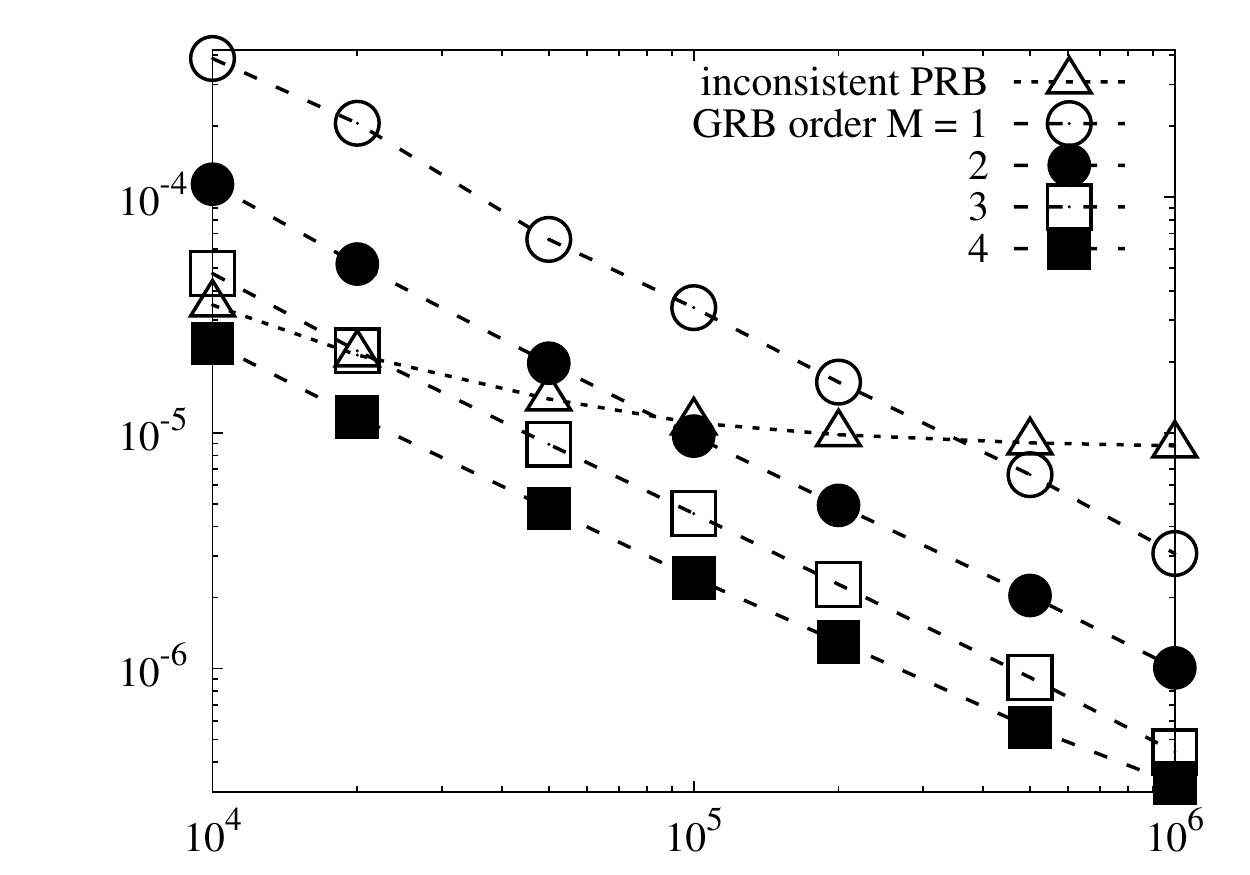} 
		\put(-255,90){\small{$C\,\|\widehat\theta-\theta^*\|_2^2$}}	
		\put(-134,-10){sample size \small{$n$}} 
	\includegraphics[width=.48\textwidth]{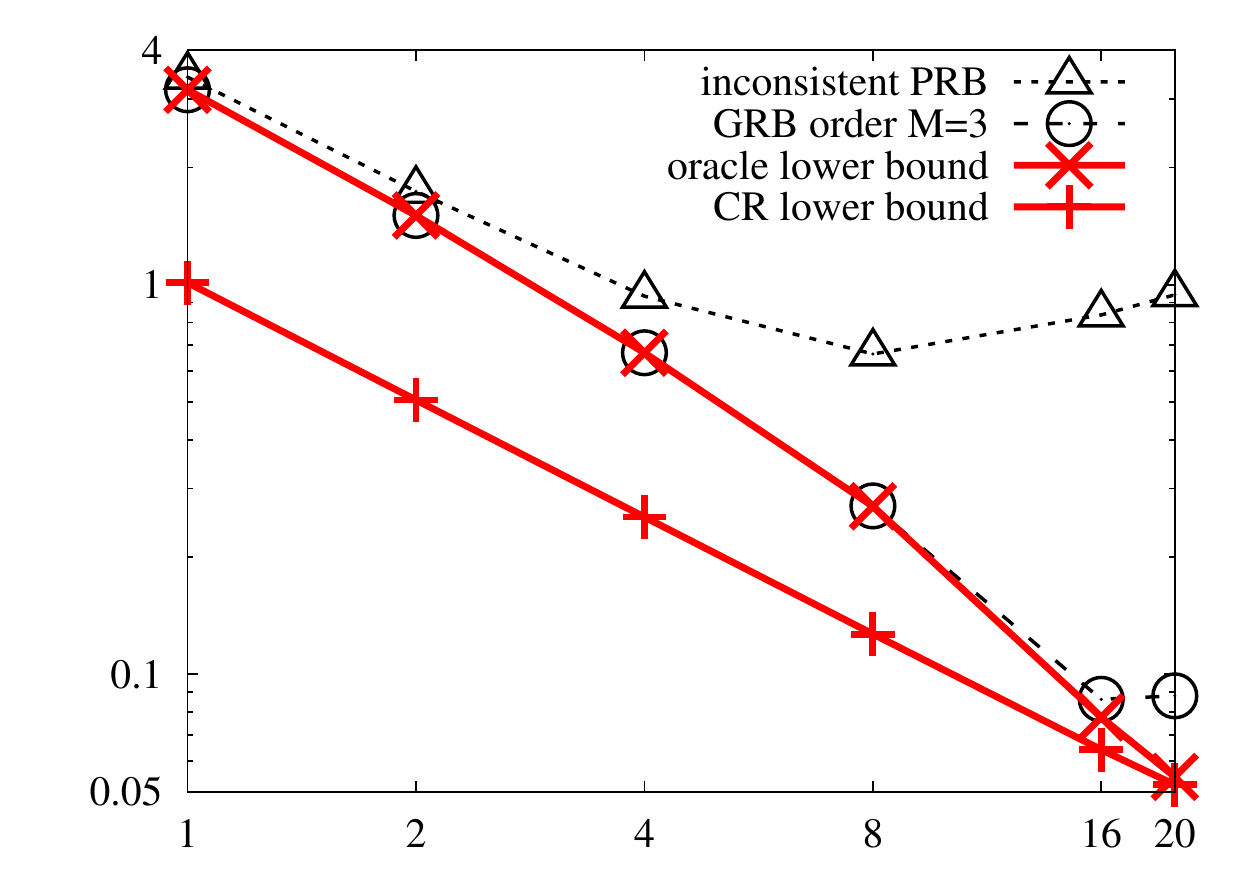}  
		\put(-152,-10){number of edges \small{$|E_j|$}}
\end{center}
\caption{ Smaller error is achieved when using more computational resources with larger $M$ and using all paired comparisons 
results in an inconsistent Pairwise Rank-Breaking (PRB)  whose error does not vanish with sample size (left). 
	Generalized Rank-Breaking (GRB) utilizes all the observations achieving the oracle lower bound (right).}
\label{fig:fig1}
\end{figure}

Figure \ref{fig:fig1} (left panel) shows the accuracy-sample tradeoff for increasing computation $M$ on the same data.
As predicted by the anlaysis, 
generalized rank-breaking (GRB) is consistent (Remark \ref{lem:consistent}) and 
the error decays at rate $(1/n)$ (Theorem \ref{thm:main_thm}), 
and decreases with increase in $M$, order of GRB. 
For comparison, we also plot the error achieved by pairwise rank-breaking (PRB) approach where we 
include all paired relations derived from data, which we call {\em inconsistent PRB}. 
As predicted by \cite{APX14a}, this results in an inconsistent estimate, whose error does not vanish as we increase the sample size. 
Notice that including all paired comparisons increases bias, but also decreases variance of the estimate. 
Hence, when sample size is limited and variance is dominating the bias, 
it is actually beneficial to include those biased paired relations to gain in variance at the cost of increased bias. 
Theoretical analysis of such a bias-variance tradeoff is outside the scope of this paper, but proposes an interesting research direction. 
We fix $d=256$, $\tilde{\ell} = 5$, $\tm_a = a$ for $a \in \{1,2,3,4,5\}$, and sample posets from the canonical scenario, except that each user is presented   $\kappa=32$ random  items. 
 The PL weights are chosen i.i.d.  $U[-2,2]$. 
 On the right panel, we let  $\tm_a = 3$ for all $a\in[\tell]$ and vary $\tilde{\ell}\in\{1,2,4,8,16\}$. 
We are providing more observations as $\tilde{\ell}$ where $|E_j|=\tilde{\ell}$. 
The proposed GRB with $M=3$ makes the full use of the given observations and achieve decreasing error, whereas 
for PRB the increased bias dominates the error. 
For comparisons, we provide the error achieved by 
an oracle estimator who knows the exact ordering among those items belonging to the top-sets and runs MLE. 
 For example, if $\tilde{\ell}=2$, 
 the GRB observes an ordering $(\{i_1,i_2,i_4,i_5,\ldots\}\prec\{i_{17},i_3,i_6\}\prec\{i_9,i_2,i_{11}\})$
 whereas the oracle estimator has extra information on the ordering among those top sets, i.e. 
        $(\{i_1,i_2,i_4,i_5,\ldots\}\prec i_{17} \prec i_3 \prec i_6 \prec i_9 \prec i_2 \prec i_{11}\})$.
Perhaps surprisingly, GRB is able to achieve a similar performance without this significant this extra information, 
unless $|E_j|$ is large. The performance degradation in large $|E_j|$ regime is precisely captured in our main analysis in Theorem \ref{thm:main_thm}.

\bigskip
\noindent
{\bf Notations.} 
We use $n$ to denote the number of users providing partial rankings, indexed by $j\in[n]$ where $[n]=\{1,2,\ldots,n\}$.
We use $d$ to denote the number of items, indexed by $i\in [d]$. 
Given rank-breaking graphs $\{G_j(S_j,E_j)\}_{j \in [n]}$ extracted from the posets $\{\cG_j\}$, we first define the order $M$ rank-breaking graphs $\{G_j^{(M)}(S_j,E^{(M)}_j)\}$, where $E^{(M)}_j$ is a subset of $E_j$ that includes only those edges $e_j \in E_j$ with $|T(e_j)|\leq M$. This  represents those edges that are included in the estimation for a choice of $M$. 
For finite sample analysis, the following quantities capture how the error depends on the topology of the data collected. 
Let $\kappa_j \equiv |S_j|$ and  $\ell_j \equiv |E_j^{(M)}|$. 
We index each edge $e_j$ in $E_j^{(M)}$ by $a \in [\ell_j]$ and 
define 
$m_{j,a} \equiv |T(e_{j,a})|$ for the $a$-th edge of the $j$-th rank-breaking graph and 
$r_{j,a} \equiv |T(e_{j,a})|+|B(e_{j,a})|$. 
Note that, we use tilde in subscript with $m_{j,a}$ and $\ell_j$ when $M$ is equal to $S_j$. That is $\tell_j$ is the number of edges in $E_j$ and $\tm_{j,a}$ is the size of the top-sets in those edges.
We let $p_j \equiv \sum_{a\in[\ell_j]} m_{j,a}$
denote the effective sample size for the observation $G_j^{(M)}$, such that 
the total effective sample size is $\sum_{j\in[n]} p_j$. 
Notice that although we do not explicitly write the dependence on $M$, all of the above quantities implicitly depend on the choice of $M$.

\section{Comparison graph}
\label{sec:topology}

The analysis of the optimization in \eqref{eq:estimate} shows  that, with high probability, 
$\cL_{\rm RB}(\theta)$ is strictly concave with $\lambda_2(H(\theta)) \leq -C_b \gamma_1 \gamma_2 \gamma_3 \lambda_2 (L)  < 0$ for all $\theta\in\Omega_b$
(Lemma \ref{lem:hess_bound}), and the gradient is also bounded with $\|\nabla \cL_{\rm RB}(\theta^*)\| \leq C'_b \gamma_2^{-1/2} (\sum_{j} p_j \log d)^{1/2} $
(Lemma \ref{lem:grad_bound}). 
the quantities  $\gamma_1$, $\gamma_2$, $\gamma_3$, and 
$\lambda_2(L)$, to be defined shortly, represent the topology of the data. 
This leads to Theorem \ref{thm:main_thm}: 
\begin{eqnarray}
	\|\htheta-\theta^*\|_2 \; \leq \; \frac{2\|\nabla\cL_{\rm RB}(\theta^*) \|}{-\lambda_2(H(\theta))} \; \leq \;
	C_b'' \frac{ \sqrt{\sum_j p_j \log d}}{\gamma_1 \gamma_2^{3/2}  \gamma_3 \lambda_2(L)}\;,
\end{eqnarray}
where  
$C_b,C_b'$, and $C_b''$ are constants that only depend on $b$, and $\lambda_2(H(\theta))$ is the second largest eigenvalue of a negative semidefinite Hessian matrix $H(\theta)$ of $\cL_{\rm RB}(\theta)$. 
Recall that $\theta^\top \vect{1}=0$ since we  restrict  our search in $\Omega_b$. 
Hence, the error depends on $\lambda_2(H(\theta))$ instead of $\lambda_1(H(\theta))$ whose corresponding eigen vector is the all-ones vector. 
We define a {\em comparison graph} $\H([d],E)$ as a weighted undirected graph with weights 
$A_{ii'} = \sum_{j\in[n]: i,i'\in S_j} p_j/(\kappa_j(\kappa_j-1)) $. 
The corresponding  graph Laplacian is defined as: 
\begin{eqnarray}\label{eq:L_comp}
L &\equiv & \sum_{j=1}^n \frac{p_j}{\kappa_j(\kappa_j-1)}\sum_{i < i' \in S_j} (e_i-e_{i'})(e_i-e_{i'})^{\top}\,.
\end{eqnarray}
It is immediate that $\lambda_1(L)=0$ with $\vect{1}$ as the eigenvector. 
There are remaining $d-1$ eigenvalues that sum to $\Tr(L) = \sum_j p_j$. 
The rescaled  $\lambda_2(L)$ and $\lambda_d(L)$ capture  the dependency on the topology:
\begin{eqnarray} \label{eq:alphabeta}
\alpha \equiv \frac{\lambda_2(L)(d-1)}{\Tr(L)}\;\;,\;\;\;\;\; \beta \equiv \frac{\Tr(L)}{\lambda_d(L)(d-1)}\,. 
\end{eqnarray}
In an ideal case where the graph is well connected, then the spectral gap of the Laplacian is large. 
This ensures all eigenvalues are of the same order and $\alpha=\beta=\Theta(1)$, resulting in  a smaller error rate. 
The concavity of $\cL_{\rm RB}(\theta)$ also depends on the following quantities. 
We discuss the role of the topology in Section \ref{sec:main}. 
Note that the quantities defined in this section implicitly depend on the choice of $M$, 
which controls the necessary computational power, via the definition of 
the rank-breaking $\{G_{j,a}\}$. 
{
We define the following quantities that control our upper bound. $\gamma_1$ incorporates asymmetry in probabilities of items being ranked at different positions depending upon their weight $\theta_i^*$. It is $1$ for $b=0$ that is when all the items have same weight, and decreases exponentially with increase in $b$. $\gamma_2$ controls the range of the size of the top-set with respect to the size of the bottom-set for which the error decays with the rate of $1/(\text{size of the top-set})$. The dependence in $\gamma_3$ and $\nu$ are due to weakness in the analysis, and ensures that the Hessian matrix is strictly negative definite. 
}

\begin{eqnarray} 
\gamma_1 &\equiv & \min_{j,a} \bigg\{\bigg(\frac{r_{j,a} - m_{j,a}}{\kappa_j}\bigg)^{2e^{2b}-2} \bigg\}, \; \;\gamma_2 \equiv  \min_{j,a} \bigg\{\bigg(\frac{r_{j,a} - m_{j,a}}{r_{j,a}}\bigg)^{2} \bigg\}\;, \text{ and }  \label{eq:gamma12}
\end{eqnarray}
\begin{eqnarray}
\gamma_3 \; \equiv \; 1- \max_{j,a} \bigg\{\frac{4e^{16b}}{\gamma_1} \frac{m_{j,a}^2r_{j,a}^2\kappa_j^2}{(r_{j,a}-m_{j,a})^5} \bigg\} \;,\;\;
\nu &\equiv& \max_{j,a} \bigg\{\frac{m_{j,a} \kappa_j^2}{(r_{j,a} - m_{j,a})^2} \bigg\}\,.
\label{eq:nu12}
\end{eqnarray}

%

\section{Main Results}
\label{sec:main} 

We present main theoretical analyses and numerical simulations confirming the theoretical predictions. 

\subsection{Upper bound on the achievable error}
\label{sec:ub}

We provide an upper bound on the error for the order-$M$ rank-breaking approach, showing  
the explicit dependence on the topology of the data. 
We assume each user provides 
a partial ranking according to his/her ordered partitions. 
Precisely, we assume that the set of offerings $S_j$, 
the number of subsets  $(\tl_j+1)$, and 
their respective sizes $(\tm_{j,1}, \ldots, \tm_{j,\tl_j})$ are {\em predetermined}.
Each user randomly draws a ranking of items from the PL model, and provides the partial ranking of the form $(\{i_6\}\prec \{i_5,i_4,i_3\} \prec \{i_2,i_1\})$ in the example in Figure \ref{fig:example}.
For a choice of $M$, the order-$M$ rank-breaking graph is extracted from this data. 
The following theorem provides an upper bound on the achieved error, 
and a proof is provided in Section \ref{sec:proof}.

\begin{theorem}\label{thm:main_thm}
Suppose there are $n$ users, $d$ items parametrized by $\theta^*\in\Omega_b$, 
and 
each user $j\in[n]$ is presented with a set of offerings $S_j\subseteq[d]$ 
and provides a partial ordering under the PL model. 
For a choice of $M\in{\mathbb Z}^+$,  if $\gamma_3>0$ and  the effective sample size $\sum_{j=1}^n p_j$ is large enough such that 
\begin{eqnarray} \label{eq:h_cond}
\sum_{j=1}^n p_j \; \geq \; \frac{2^{14}e^{20b}\nu^2}{(\alpha\gamma_1\gamma_2\gamma_3)^2\beta}\frac{p_{\max}}{\kappa_{\min}} d\log d\;,
\end{eqnarray}
where $b\equiv \max_i|\theta^*_i|$ is the dynamic range, $p_{\rm max}=\max_{j\in[n]} p_j$, $\kappa_{\rm min}=\min_{j\in[n]}\kappa_j$, 
$\alpha$ is the (rescaled) spectral gap, 
$\beta$ is the (rescaled) spectral radius in \eqref{eq:alphabeta}, 
and  $\gamma_1$, $\gamma_2$, $\gamma_3$, and $\nu$ are defined in  \eqref{eq:gamma12} and \eqref{eq:nu12},
then the generalized rank-breaking estimator in \eqref{eq:estimate} achieves
\begin{eqnarray}
\frac{1}{\sqrt{d}}\norm{\widehat{\theta} - \theta^*} & \leq & 
\frac{40 e^{7b} }{\alpha\gamma_1\gamma_2^{3/2} \gamma_3 }\sqrt{\frac{d\log d}{\sum_{j=1}^n \sum_{a =1}^{\ell_j} m_{j,a}}}\,,
\label{eq:main}
\end{eqnarray}
with probability at least $1-3e^{3}d^{-3}$. 
Moreover, for $M \leq 3$ the above bound holds with  $\gamma_3$ replaced by one, giving a tighter result. 
\end{theorem}
Note that the dependence on the choice of  $M$ is not explicit in the bound, but rather is implicit in the 
construction of the comparison graph and the number of effective samples 
$N = \sum_{j} \sum_{a \in [\ell_j]} m_{j,a} $.
In an ideal case, $b=O(1)$ and $m_{j,a}=O(r_{j,a}^{1/2})$ for all $(j,a)$ 
such that $\gamma_1,\gamma_2$ are finite. 
further, if the spectral gap is large such that $\alpha>0$ and $\beta>0$, 
then Equation  \eqref{eq:main} implies that we need the effective sample size to scale as $O(d\log d)$, 
which is only a logarithmic factor larger than the number of parameters. 
In this ideal case, 
 there exist universal constants $C_1,C_2$ such that 
if $m_{j,a} < C_1\sqrt{r_{j,a}}$ and $r_{j,a} > C_2 \kappa_j$ for all $\{j,a\}$, then 
the condition $\gamma_3>0$  is met. 
Further, when $r_{j,a}=O(\kappa_{j,a})$, $\max \kappa_{j,a}/\kappa_{j',a'} = O(1)$, 
and $\max p_{j,a}/p_{j',a'}=O(1)$, then condition on the effective sample size is met with $\sum_j p_j = O(d \log d)$. 
We believe that dependence in $\gamma_3$  is weakness of our analysis and there is no dependence as long as  $m_{j,a} < r_{j,a}$. 

\subsection{Lower bound on computationally unbounded estimators}
\label{sec:lb}
Recall that  $\tilde{\ell}_j \equiv |E_j|$, $\tilde{m}_{j,a}=|T(e_a)|$ and $\tilde{r}_{j,a}=|T(e_a)\cup B(e_a)|$ when $M = S_j$. 
We prove a fundamental lower bound on the achievable error rate that holds for 
any {\em unbiased} estimator even with no restrictions on the computational complexity. 
For each $(j,a)$, define $\eta_{j,a}$ as 
\begin{align}
	\label{eq:eta}
	&\eta_{j,a}  =  \sum_{u=0}^{\tm_{j,a}-1} \Big( \frac{1}{\tr_{j,a}-u} + 
	\frac{u(\tm_{j,a}-u)}{\tm_{j,a}(\tr_{j,a}-u)^2} \Big) + 
	 \sum_{u<u'\in[\tm_{j,a}-1]} \frac{2u}{\tm_{j,a}(\tr_{j,a}-u)} \frac{\tm_{j,a}-u'}{\tr_{j,a}-u'} \\
	& \;\;\;\; \;=  \;  \tm_{j,a}^2/(3\tr_{j,a} ) + O(\tm_{j,a}^3/\tr_{j,a}^2) \,.
\end{align}

\begin{theorem} \label{thm:lower_bound}
	Let $\U$ denote the set of all unbiased estimators of $\theta^*$ that are centered such that $\htheta\vect{1}=0$, 
	and let $\mu=\max_{j\in[n],a\in[\tell_j]}\{\tm_{j,a}-\eta_{j,a}\}$. 
	For all $b>0$, 
	\begin{eqnarray}
	\inf_{\widehat{\theta} \in \U} \sup_{\theta^* \in \Omega_b} \E[\norm{\widehat{\theta}-\theta^*}^2] &\geq & \max \left\{  \frac{(d-1)^2}{\sum_{j=1}^n\sum_{a=1}^{\tell_j}(\tm_{j,a}-\eta_{j,a})} \;,\; \frac{1}{\mu}\sum_{i=2}^d \frac{1}{\lambda_i(L)} \right\} \,.
	\label{eq:lb}
	\end{eqnarray}
\end{theorem}
The proof relies on the Cramer-Rao  bound and is provided in Section \ref{sec:proof}. 
Since $\eta_{j,a}$'s are non-negative, the mean squared error is lower bounded by $(d-1)^2/N$, where 
$N=\sum_{j} \sum_{a \in \tell_j} \tm_{j,a}$ is the effective sample size. 
Comparing it to the upper bound in \eqref{eq:main}, this  is tight 
up to a logarithmic factor when 
$(a)$ the topology of the data is well-behaved such that all respective quantities are finite; and 
$(b)$ there is no limit on the computational power and $M$ can be made as large as we need. 
The bound in Eq.  \eqref{eq:lb} further gives  a tighter lower bound,  
capturing the dependency in $\eta_{j,a}$'s and $\lambda_i(L)$'s. 
Considering the first term, $\eta_{j,a}$ is larger when $\tm_{j,a}$ is close to $\tr_{j,a}$, giving a tighter bound. 
The second term in \eqref{eq:lb} implies we get a tighter bound when $\lambda_2(L)$ is smaller. 


\begin{figure}[h]
 \begin{center}
	\includegraphics[width=.32\textwidth]{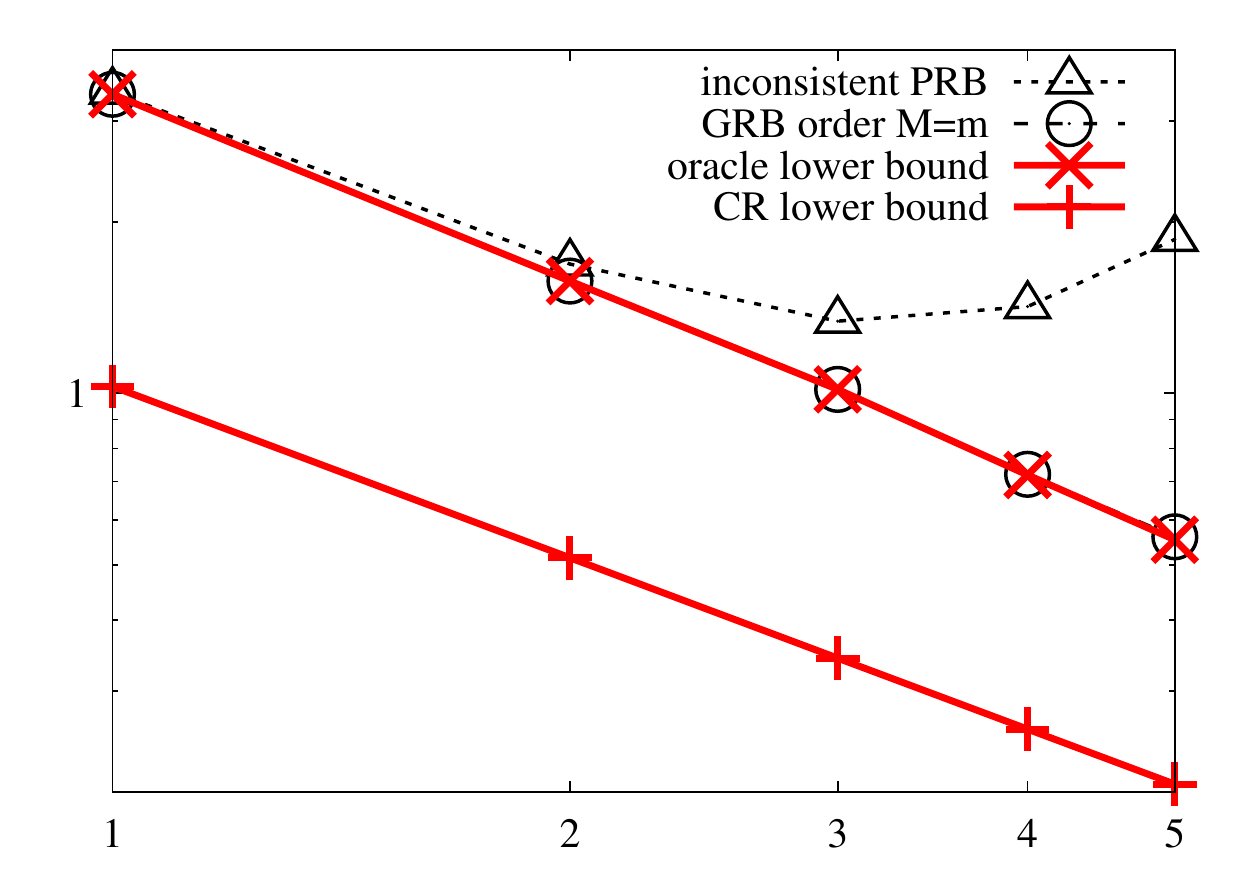} 
	\put(-199,55){\small{$C\,\|\widehat\theta-\theta^*\|_2^2$}}	
		\put(-110,-7){size of top-set \small{$m$}} 
	\includegraphics[width=.32\textwidth]{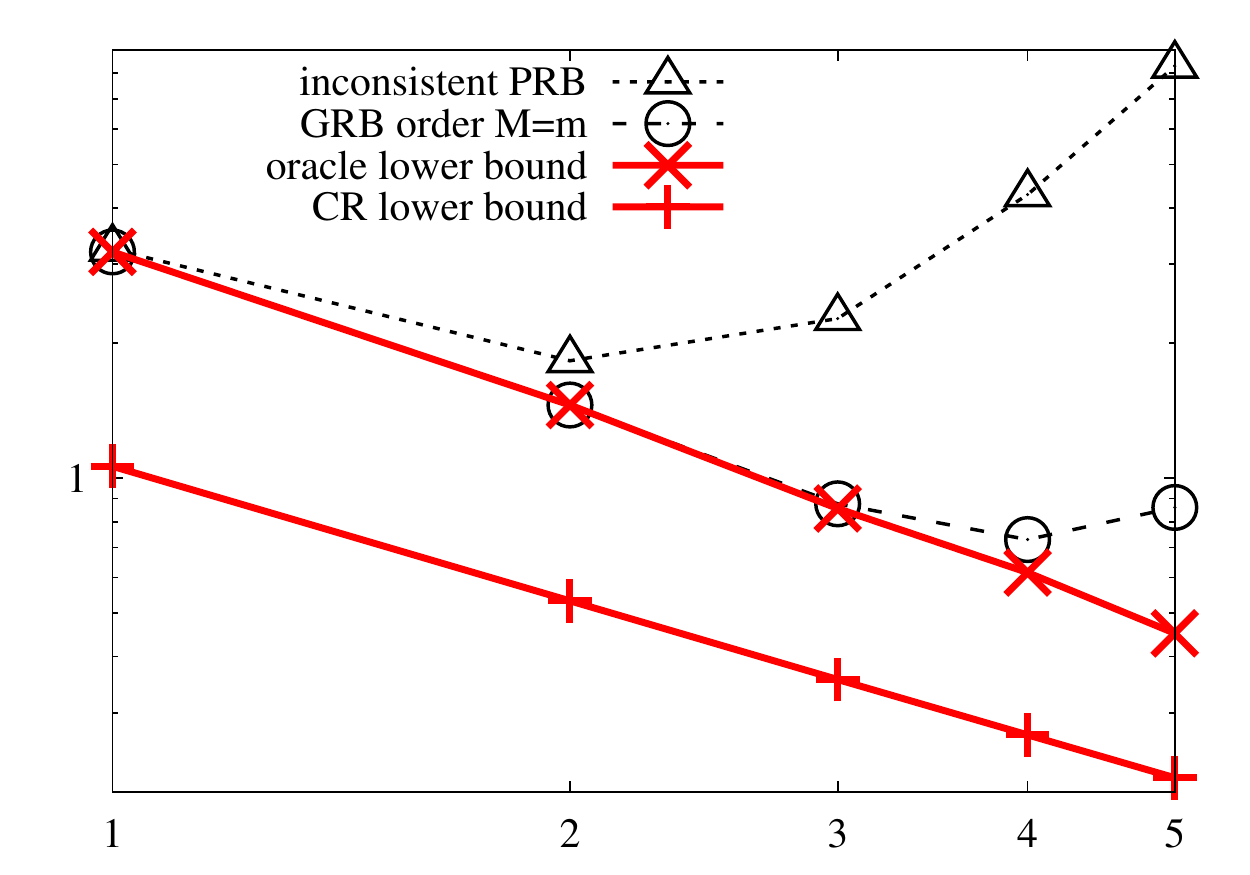}  
	\put(-110,-7){size of top-set \small{$m$}}
   \includegraphics[width=.32\textwidth]{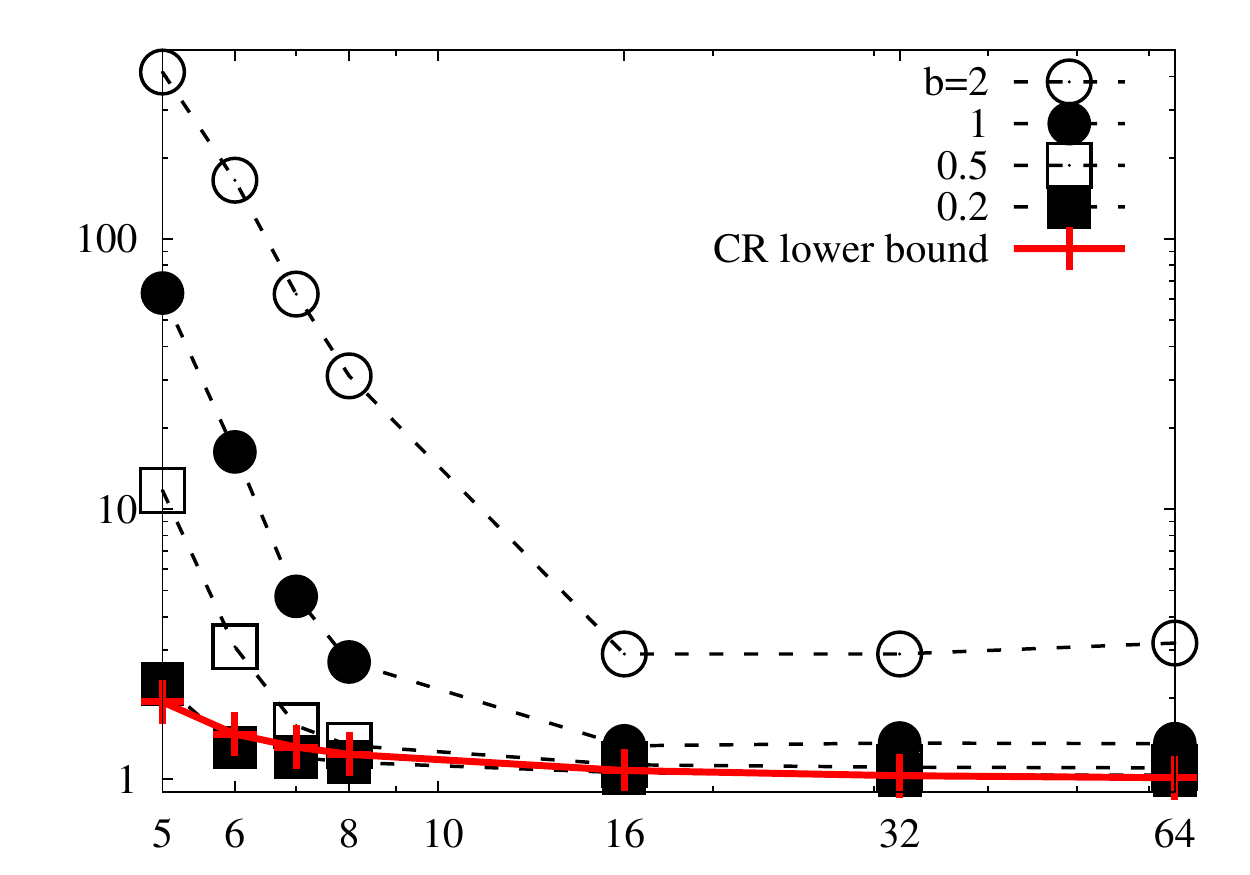}
	\put(-92,-7){set-size $\kappa$}	
\end{center}
\caption{Accuracy degrades as   $(\kappa-m)$ gets small and as the dynamic range $b$ gets large.}
\label{fig:fig2}
\end{figure}

In Figure \ref{fig:fig2} left and middle panel, 
we compare performance of our algorithm with pairwise breaking, Cramer Rao lower bound and oracle MLE lower bound.
We fix $d=512$, $n= 10^5$, $\theta^*$ chosen i.i.d. uniformly over $[-2,2]$. 
 Oracle MLE knows relative ordering of items in all the top-sets $T(e)$ and hence is strictly better than the GRB. 
We fix $\tell=\ell=1$ that is $r = \kappa$, and vary $m$ .  
 In the left panel, we fix $\kappa=32$ and in the middle panel, we fix $\kappa=16$.  
 Perhaps surprisingly, GRB matches with the oracle MLE which means relative ordering of top-$m$ items among themselves is statistically insignificant when $m$ is sufficiently small in comparison to $\kappa$.  
 For $\kappa=16$, as $m$ gets large, the error starts to increase as predicted by our analysis.  
 The reason is that 
 the quantities $\gamma_1$ and $\gamma_2$ gets smaller as  $m$ increases, and the upper bound increases consequently. 
 In the right panel, we fix $m=4$. 
  When $\kappa$ is small,  $\gamma_2$ is small, and hence error is large;
 when $b$ is large $\gamma_1$ is exponentially small, and hence error is significantly large. 
This is different from learning Mallows models in \cite{AM12} where peaked distributions are easier to learn, 
and is related to the fact that we are not only interested in recovering the (ordinal) ranking but also the (cardinal) weight. 
 

\subsection{ Computational and statistical tradeoff }

For estimators with limited   computational power, however,  
the above lower bound fails to capture the dependency on the allowed computational power. 
Understanding such fundamental trade-offs is 
a challenging problem, which has been studied only in a few special cases, e.g. planted clique problem \citep{DM15,MPW15}. 
This is outside the scope of this paper, and 
we instead  investigate the trade-off achieved by the proposed rank-breaking approach. 
When we are limited on computational power, 
Theorem \ref{thm:main_thm}  
implicitly captures this dependence when 
order-$M$ rank-breaking is used. 
The dependence is captured indirectly via the resulting rank-breaking $\{G_{j,a}\}_{j\in[n],a\in[\ell_j]}$
and the topology of it.
We make this trade-off explicit by considering a simple but  canonical example. 
Suppose $\theta^*\in\Omega_b$ with $b=O(1)$. 
Each user gives an i.i.d. partial ranking, where all items are offered and the partial ranking 
is based on an ordered partition with $\tilde{\ell}_j = \lfloor\sqrt{2c}d^{1/4}\rfloor$ subsets. The top subset has size $\tilde{m}_{j,1}=1$, and the $a$-th subset has  size $\tilde{m}_{j,a}=a$, up to $a < \tilde{\ell}_j$, in order to ensure that they sum at most to $c\sqrt{d}$ for sufficiently small positive constant $c$ and the condition on $\gamma_3 >0$ is satisfied. 
The last subset includes all the remaining items in the bottom, ensuring $\tilde{m}_{j,\tilde{\ell}_j} \geq d/2$ and  $\gamma_1,\gamma_2$ and $\nu$ are all finite.  

{\bf Computation.} For a choice of $M$ such that $M\leq \ell_j-1$, 
we consider the computational complexity in evaluating the gradient of $\cL_{\rm RB}$, 
which scales as $T_{M}  = \sum_{j\in[n]} \sum_{a\in[M]} (m_{j,a}!)r_{j,a} = O(M!\times dn )$. 
Note that  we find the MLE by solving a convex optimization problem using first order methods, and 
 detailed analysis of the convergence rate and the complexity of solving general convex optimizations is outside the scope of this paper. 

{\bf Sample.} Under the canonical setting, for $M\leq \ell_j-1$, we have 
$L = M(M+1)/(2d(d-1))  \big({\mathbb I} - \vect{1}\vect{1}^\top\big)$.  
This complete graph has the largest possible spectral gap, and hence $\alpha>0$ and $\beta>0$.
Since the effective samples size is $\sum_{j,a}\tilde{m}_{j,a}\I\{\tilde{m}_{j,a}\leq M\} = n M(M+1)/2$, 
it follows from Theorem \ref{thm:main_thm} that the (rescaled) root mean squared error is  
$O(\sqrt{(d\log d )/ (n M^2)})$.
In order to achieve a 
target error rate of $\varepsilon$, we need to choose $M=\Omega((1/\varepsilon)\sqrt{(d\log d)/n})$. 
The resulting trade-off between run-time and sample to achieve root mean squared error $\varepsilon$ is $T(n) \propto (\lceil (1/\varepsilon)\sqrt{(d\log d) / n}  \rceil)! d n $. 
We show numerical experiment under this canonical setting in Figure \ref{fig:fig1} (left) 
with $d=256$ and $M\in\{1,2,3,4,5\}$, illustrating the trade-off in practice.

%
%
%

\subsection{Real-world datasets}

\begin{figure}[h]
 \begin{center}
	\includegraphics[width=.32\textwidth]{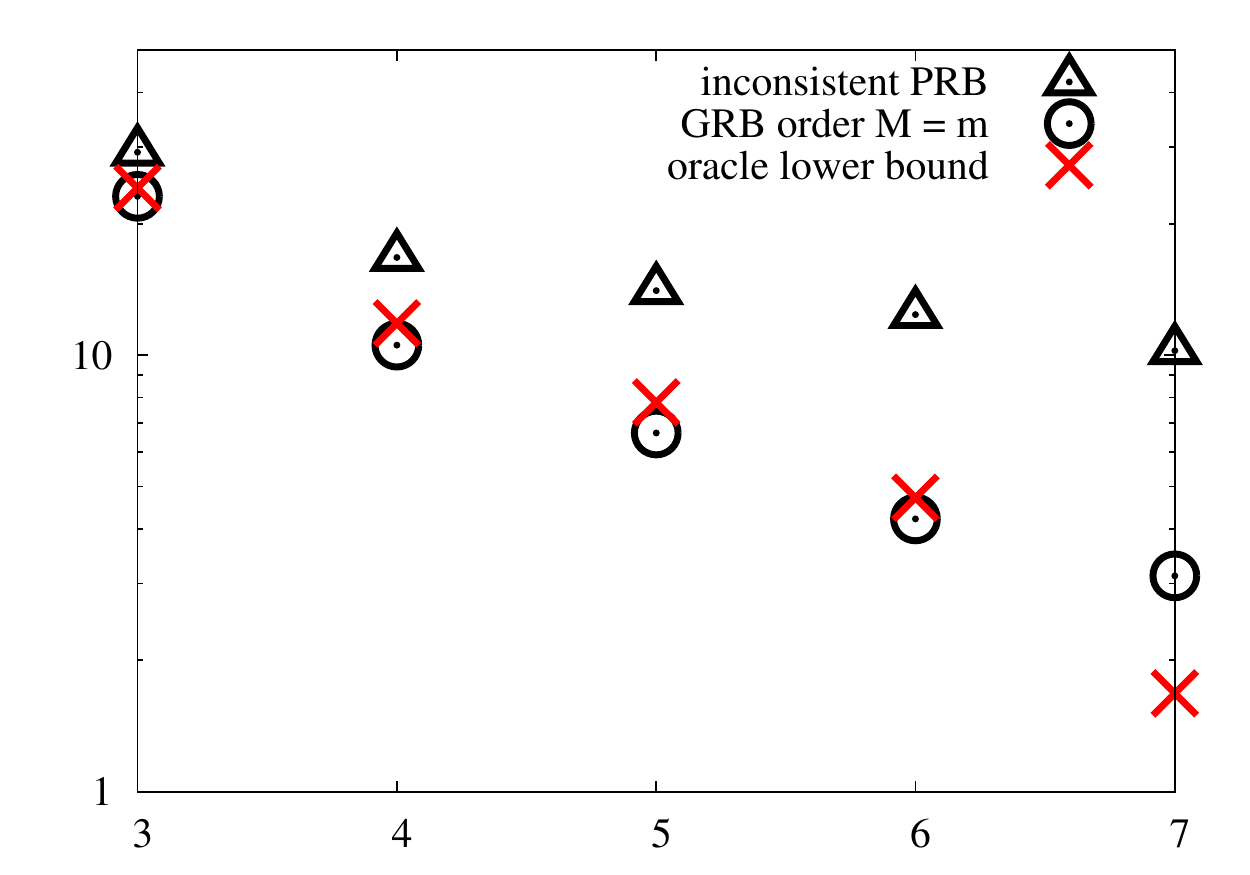} 
		\put(-192,55){\small{$C\,\|\widehat\theta-\theta^*\|_2^2$}}	
		\put(-106,-7){size of top-set \small{$m$}} 
	\includegraphics[width=.32\textwidth]{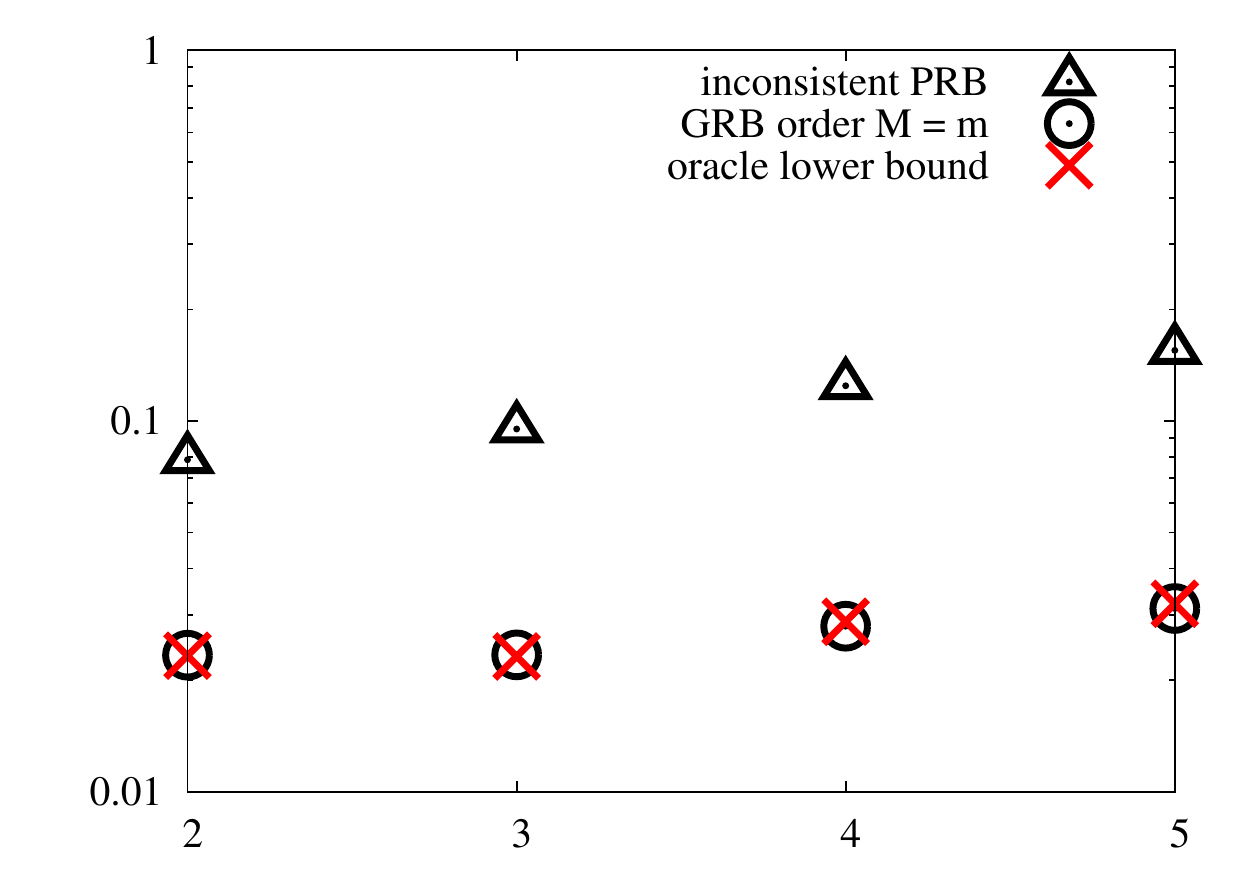}  
		\put(-104,-7){size of top-sets \small{$m$}}
   \includegraphics[width=.32\textwidth]{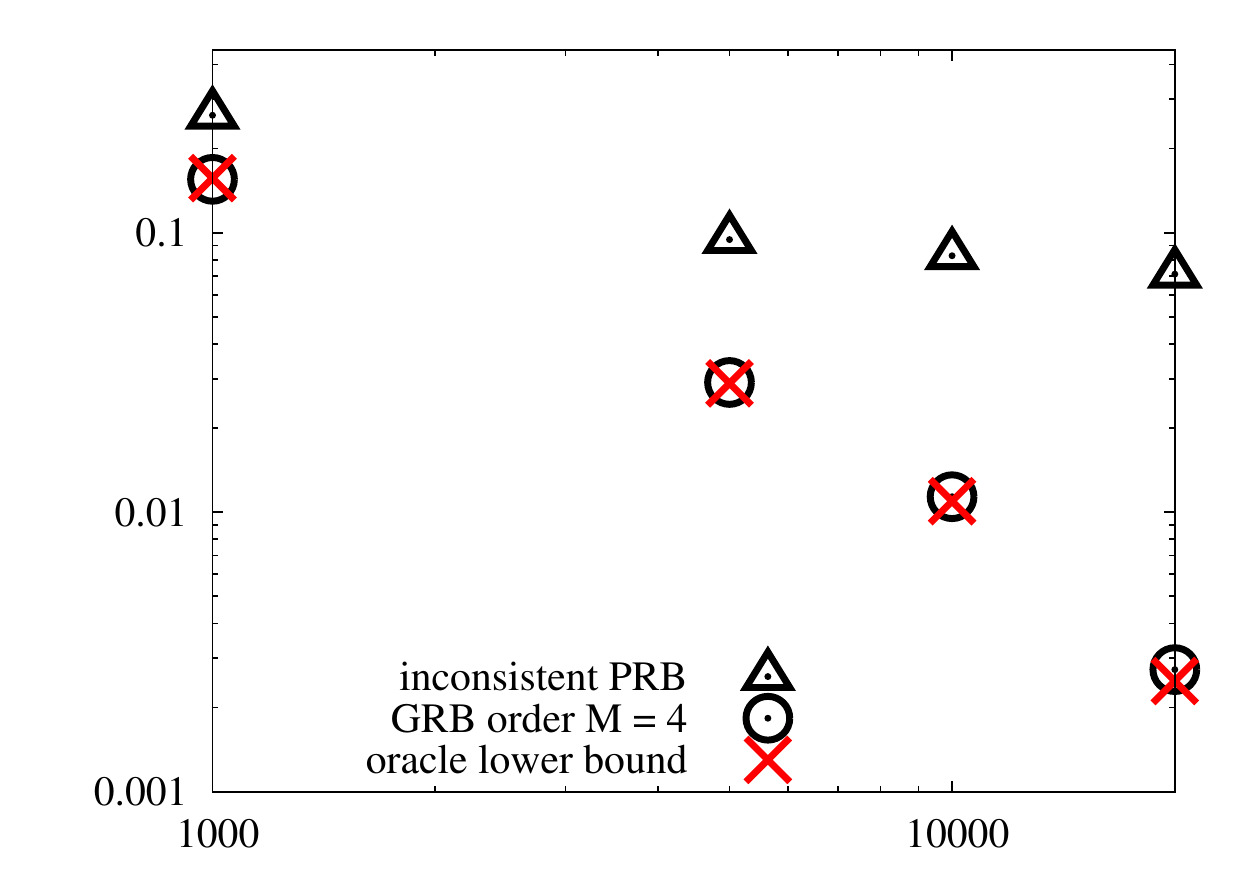}
		\put(-95,-7){sample size $n$}	
\end{center}
\caption{Generalized rank-breaking improves over pairwise RB and is close to  oracle MLE.}
\label{fig:fig3}
\end{figure}

On sushi preferences \citep{Kam03} and jester dataset \citep{GRG01}, we  improve over pairwise breaking and achieves same performance as the oracle MLE. 
 Full rankings over $\kappa = 10$ types of sushi are randomly chosen from $d=100$ types of sushi are provided by $n= 5000$ individuals. 
 As the ground truth $\theta^*$, we use the ML estimate of PL weights over the entire data. 
 In Figure \ref{fig:fig3}, left panel, for each $m \in \{3,4,5,6,7\}$, we remove the known ordering among the top-$m$ and bottom-$(10-m)$ sushi in each set, and run our estimator with one breaking edge between top-$m$ and bottom-$(10-m)$ items. 
 We compare our algorithm with inconsistent pairwise breaking (using optimal choice of parameters from \cite{KO16}) and the oracle MLE. 
 For $m\leq 6$, the proposed rank-breaking performs as well as an oracle who knows the hidden ranking among the top $m$ items. 
 Jester dataset consists of continuous ratings between $-10$ to $+10$ of $100$ jokes on sets of size $\kappa$,  $36 \leq \kappa \leq 100$, by $24,983$ users. 
 We convert ratings into full rankings. The ground truth $\theta^*$ is computed similarly. 
 For $m \in \{2,3,4,5\}$, we convert each full ranking into a poset that has $\ell= \lfloor\kappa/m\rfloor$ partitions of size $m$, by removing known relative ordering from each partition. Figure \ref{fig:fig3} compares the three algorithms using all  samples (middle panel), and by varying the sample size (right panel) for fixed $m=4$. All figures are averaged over $50$ instances.

%
%
%

\section{Proofs}
\label{sec:proof}
We provide the proofs of the main results. 

\subsection{Proof of Remark \ref{lem:concave}}
\label{sec:concave}

Recall that $\P_{\theta}( B(e) \prec T(e) )$ is the probability that 
an agent ranks the collection of items $T(e)$ above $B(e)$ when offered $S = B(e)\cup T(e)$. 
 We want to show that $\P_{\theta}( B(e) \prec T(e) )$ is log-concave under the PL model. 
 We prove a slightly general result which works for a family of RUMs in the location family. 
Random Utility Models (RUM) are defined as a probabilistic model where 
there is a real-valued utility parameter $\theta_i$ associated with each items $i \in S$, and 
an agent  independently samples random utilities $\{U_i\}_{i \in S}$ for each item $i$ with conditional distribution $\mu_{i}(\cdot|\theta_i)$. Then the ranking is obtained by sorting the items in decreasing order as per the observed random utilities $U_i$'s. 
{\em{Location family}} is a subset of RUMs where the shapes of $\mu_i$'s are fixed and the only parameters are the means of the distributions. For location family, the noisy utilities can be written as $U_i = \theta_i + Z_i$ for i.i.d. random variable $Z_i$'s. In particular, it is PL model when $Z_i$'s follow the independent standard Gumbel distribution. 
We will show that for the location family if the probability density function for each $Z_i$'s is log-concave then $\log\P_{\theta}( B(e) \prec T(e) )$ is concave. 
The desired claim  follows as the pdf of standard Gumbel distribution is log-concave. We use the following Theorem from \cite{Pre80}. 
A similar technique was used to prove concavity when $|T(e)|=1$ in \cite{APX12}. 
\begin{lemma}[Theorem 9 in \cite{Pre80}]  
 
Suppose $g(\theta,Z)$ is a concave function in $\reals^{2r}$, where $\theta\in \reals^r$ is fixed and $Z$ is a $r-$component random vector whose probability distribution is logarithmic concave in $\reals^r$, then the function
\begin{eqnarray}
h(\theta) = \P[g(\theta,Z) \geq 0], \qquad \text{ for } \theta \in \reals^r
\end{eqnarray}  
is logarithmic concave on $\reals^r$.
\end{lemma}
To apply the above lemma to get our result, let $r = | S |$, 
$g(\theta,Z) = \min_{i \in T(e) } \{\theta_i+ Z_i\} - \max_{i' \in B(e)} \{\theta_{i'}+ Z_{i'}\}$, and observe that $\P_{\theta}(B(e)\prec T(e)) = \P(g(\theta,Z) \geq 0)$ and $g(\theta,Z)$ is concave. 

\subsection{Proof of Remark \ref{lem:consistent} }
\label{sec:consistent}
Define event $E(e) \equiv \{ T(e)\cup B(e) \text{ items are ranked in bottom $r$ positions when the offer set is $[d]$}\}$.  Define $\P_{\theta,[d]}(B(e) \prec T(e) | E(e))$ be the conditional probability of $T(e)$ items being ranked higher than $B(e)$ items when the offer set is $[d]$, conditioned on the event $E(e)$. Observe that $\P_{\theta,[d]}(B(e) \prec T(e) | E(e))$ is the probability of observing the event $B(e) \prec T(e)$ under the proposed rank-breaking. First we show that $\P_{\theta}(e) = \P_{\theta,[d]}(B(e) \prec T(e) | E(e))$, where $\P_{\theta}(e)$ is the probability that $T(e)\prec B(e)$ when the offer set is $\{T(e)\cup B(e)\}$ as defined in \eqref{eq:pair}. This follows from the fact that under PL model for any disjoint set of items $\{\C_i\}_{i \in [{\ell}]}$ such that $\cup_{i=1}^{\ell}\C_i = [d]$, 
\begin{align}
\P\big(\C_\ell \prec \C_{\ell-1} \prec \cdots \prec \C_1\big) = \P\big(\C_{\ell} \prec \C_{\ell-1}\big) \P\big(\{\C_{\ell},\C_{\ell-1}\} \prec \C_{\ell-2}\big)\cdots\P\big(\{\C_{\ell},\C_{\ell-1},\cdots,\C_2\} \prec \C_{1}\big)\,,
\end{align}
where $\prob(\C_{i_1}\prec \C_{i_2})$ is the probability that $\C_{i_2}$ items are ranked higher than $\C_{i_1}$ items when the offer set is $S = \{\C_{i_1}\cup \C_{i_2}\}$.
Under the given sampling scenario, the comparison graph $\H([d],E)$ as defined in section \ref{sec:topology} is connected and hence the estimate $\widehat{\theta}$, \eqref{eq:estimate} is unique. Therefore, it follows that maximum likelihood estimate $\widehat{\theta}$ is consistent. Further, for a general sampling scenario, Theorem \ref{thm:main_thm} proves that the estimator is consistent as the error goes to zero in the limit as $n$ increases.

\subsection{Proof of Theorem \ref{thm:main_thm}}
\textcolor{black}
{We define few additional notations. $ p \equiv (1/n)\sum_{j=1}^n p_j$. $V(e_{j,a}) \equiv T(e_{j,a}) \cup B(e_{j,a})$ for all $j \in [n]$ and $a \in [\ell_j]$. Note that by definition of rank-breaking edge $e_{j,a}$, $V(e_{j,a})$ is a random set of items that are ranked in bottom $r_{j,a}$ positions in a set of $S_j$ items by the user $j$.}
 
The proof sketch is inspired from \cite{KO16}. 
The main difference and technical challenge is in showing the strict concavity of $\L_{\RB}(\theta)$ when restricted to $\Omega_b$. 
We want to prove an upper bound on  $\Delta = \widehat{\theta} - \theta^*$, 
where $\htheta$ is the sample dependent solution of the optimization \eqref{eq:estimate} and 
$\theta^*$ is the true utility parameter from which the samples are drawn.
Since $\widehat{\theta}, \theta^* \in \Omega_b$, it follows that $\Delta \vect{1} = 0$. Since $\widehat{\theta}$ is the maximizer of $\L_{\RB}(\theta)$, we have the following inequality,
\begin{eqnarray}\label{eq:main1}
\Lrb(\widehat{\theta}) - \Lrb(\theta^*) - \langle\nabla\Lrb(\theta^*),\Delta\rangle \;\geq\; -\langle\nabla\Lrb(\theta^*),\Delta\rangle \;\geq\; -\norm{\nabla\Lrb(\theta^*)}_2\norm{\Delta}_2, 
\end{eqnarray} 
where the last inequality uses the Cauchy-Schwartz inequality. By the mean value theorem, 
there exists a $\theta = c\widehat{\theta} + (1-c)\theta^*$ for some $c \in [0,1]$ such that $\theta \in \Omega_b$ and 
\begin{eqnarray}\label{eq:main2}
\Lrb(\widehat{\theta}) - \Lrb(\theta^*) - \langle\nabla\Lrb(\theta^*),\Delta\rangle \; =\; \frac{1}{2}\Delta^\top H(\theta)\Delta \leq -\frac{1}{2}\lambda_2(-H(\theta))\norm{\Delta}_2^2,
\end{eqnarray}  
where $\lambda_2(-H(\theta))$ is the second smallest eigen value of $-H(\theta)$. 
We will show in Lemma \ref{lem:hess_bound} that 
$-H(\theta)$ is positive semi definite with one eigen value at zero with a corresponding eigen vector $\vect{1}=[1,\ldots,1]^\top $. 
The last inequality follows since $\Delta^\top\vect{1} = 0$. 
Combining Equations \eqref{eq:main1} and \eqref{eq:main2},
\begin{eqnarray} \label{eq:main3}
\norm{\Delta}_2 \;\;\leq\;\; \frac{2\norm{\nabla\Lrb(\theta^*)}_2}{\lambda_2(-H(\theta))}, 
\end{eqnarray}  
where we used the fact that $\lambda_2(-H(\theta))>0$ from Lemma \ref{lem:hess_bound}. 
The following technical lemmas prove 
that the norm of the gradient is upper bounded by 
$ \gamma_2^{-1/2}e^{b} \sqrt{6 np\log d}$ 
with high probability 
and the second smallest eigen value is lower bounded by 
$(1/8)\,e^{-6b}\alpha\gamma_1\gamma_2\gamma_3(np/(d-1))$. 
This finishes the proof of Theorem \ref{thm:main_thm}.

The (random) gradient of the log likelihood in \eqref{eq:estimate} can be written as the following, where the randomness is in 
which items ended up in  the top set $T(e_{j,a})$ and the bottom set $B(e_{j,a})$: 
\begin{eqnarray} \label{eq:grad_def}
\nabla_i \L_{\RB}(\theta) &=& \sum_{j = 1}^n \sum_{a = 1}^{\ell_j} \sum_{\substack{\C \subseteq S_j,\\ |\C|=r_{j,a}-1}} \I\big\{\, V(e_{j,a})=  \{\C,i\} \,\big\}  \frac{\partial \log \P_{\theta}(e_{j,a})}{\partial \theta_i}\,.
\end{eqnarray}
Note that we are intentionally decomposing each summand  as a summation over 
all $\C$ of size $r_{j,a}-1$, such that we can separate the analysis of the expectation in the following lemma. 
The random variable $\I\{ \{\C,i\} = V(e_{j,a})\}$ indicates that we only include one term for any given instance of the sample. 
\textcolor{black}
{Note that the event $\I\{ \{\C,i\} = V(e_{j,a})\}$ is equivalent to the event that the $\{\C,i\}$ items are ranked in bottom $r_{j,a}$ positions in the set $S_j$, that is $V(e_{j,a})$ items are ranked in bottom $r_{j,a}$ positions in the set $S_j$.}  
\begin{lemma}\label{lem:grad_expec}
	If the $j$-th poset is drawn  from the PL model with weights $\theta^*$ 
	then for any given $\C' \subseteq S_j$ with $|\C'|=r_{j,a}$, 
	\begin{eqnarray}
	 \E\bigg[\I\big\{\C' = V(e_{j,a}) \big\}  \frac{\partial \log \P_{\theta^*}(e_{j,a})}{\partial \theta^*_i}  \bigg| \{e_{j,a'}\}_{a' < a}\bigg] &=& 0\,.
	\end{eqnarray}
\end{lemma}
First, this lemma implies that 
$\E\big[\I\big\{\,  \C' = V(e_{j,a}) \,\big\}  \frac{\partial \log \P_{\theta^*}(e_{j,a})}{\partial \theta^*_i}\big] = 0$. 
Secondly, the above lemma allows us to construct a vector-valued martingale and 
apply a generalization of Azuma-Hoeffding's tail bound on the norm to prove the following concentration of measure.
This proves the desired bound on the gradient. 
\begin{lemma}\label{lem:grad_bound}
	If $n$ posets are independently drawn over $d$ items  from the PL model with weights $\theta^*$ 
	then 
	with probability at least $1 - 2e^3d^{-3}$, 
	\begin{eqnarray}
	\norm{\nabla \L_{\RB}(\theta^*) } \leq \gamma_2^{-1/2}e^{b} \sqrt{6np\log d }\,,
	\end{eqnarray}
	where $\gamma_2$ depend on the choice of the rank-breaking and are defined in Section \ref{sec:topology}. 
\end{lemma}

We will prove in \eqref{eq:defHessian} that the Hessian matrix $H(\theta) \in \cS^d$ with $H_{ii'}(\theta) = \frac{\partial^2 \L_{\RB}(\theta)}{\partial\theta_i\partial\theta_{i'}}$ can be expressed as 
\begin{eqnarray} \label{eq:hess_def}
-H(\theta) = \sum_{j=1}^n\sum_{a= 1}^{\ell_j} \sum_{i < i' \in S_j} \I\{(i,i') \subseteq V(e_{j,a})\} \left(  \;\frac{\partial^2 \log \P_{\theta}(e_{j,a})}{\partial\theta_i\partial\theta_{i'}} (e_i - e_{i'})(e_i - e_{i'})^\top\right)\,.
\end{eqnarray}
It is easy to see that $H(\theta) \vect{1}=0$.
The following lemma proves a lower bound on the second smallest eigenvalue 
$\lambda_{2}(-H(\theta))$ in terms of  re-scaled spectral gap $\alpha$ of the comparison graph $\H$ 
defined in Section  \ref{sec:topology}.
\begin{lemma} \label{lem:hess_bound}
Under the hypothesis of Theorem \ref{thm:main_thm}, 
if the assumptions in Equation \eqref{eq:h_cond} are satisfied then with probability at least $1-d^{-3}$, the following holds for any $\theta \in \Omega_b$:
\begin{eqnarray} \label{eq:hess_bound}
\lambda_2(-H(\theta)) & \geq & \frac{e^{-6b}\alpha\gamma_1\gamma_2\gamma_3}{8}\frac{np}{(d-1)}\,,
\end{eqnarray}
and  $\lambda_1(-H(\theta))=0$ with corresponding eigen vector $\vect{1}$. 
\end{lemma}
This finishes the proof of the desired claim.

\subsection{Proof of Lemma \ref{lem:grad_expec}}
\textcolor{black}
{Recall that $e_{j,a}$ is a random event where randomness is in which items ended up in the top-set $T(e_{j,a})$ and the bottom-set $B(e_{j,a})$, and $\P_{\theta^*}(e_{j,a}) = \P_{\theta^*}[B(e_{j,a}) \prec T(e_{j,a})]$ that is the probability of observing $B(e_{j,a}) \prec T(e_{j,a})$ when the offer set is $B(e_{j,a}) \cup T(e_{j,a})$ as defined in \eqref{eq:pair}. Define, $\P_{\theta^*,S_j}[e_{j,a}| V(e_{j,a}) = \C' ]$ to be the conditional probability of observing $B(e_{j,a}) \prec T(e_{j,a})$, when the offer set is $S_j$, conditioned on the event that $V(e_{j,a}) = \C' $. Note that we have put subscript $S_j$ in $\P_{\theta^*}$ to specify that the offer set is $S_j$. Observe that for any set $\C' \subseteq S_j$, the event $\{\C' = V(e_{j,a})\}$ is equivalent to $\C'$ items being ranked in bottom $r_{j,a}$ positions when the offer set is $S_j$. In other words, it is conditioned on the event that the subset $V(e_{j,a})$ items are ranked in bottom $r_{j,a}$ positions when the offer set is $S_j$. It is easy to check that under PL model 
$$\P_{\theta^*,S_j}[e_{j,a} | V(e_{j,a}) = \C'] = \P_{\theta^*}[e_{j,a}],$$  
(see Remark \ref{lem:consistent}). Also, by conditioning on any outcome of $\{e_{j,a'}\}_{a' < a}$ it can be checked that 
$$\P_{\theta^*,S_j}[e_{j,a} | V(e_{j,a}) = \C', \{e_{j,a'}\}_{a' < a}]= \P_{\theta^*,S_j}[e_{j,a}| V(e_{j,a}) = \C'].$$ 
Therefore, we have
\begin{align}
&\E\bigg[ \frac{\partial \log \P_{\theta^*}\big[e_{j,a}\big]}{\partial \theta^*_i} \bigg|  V(e_{j,a})=\C', \{e_{j,a'}\}_{a' < a} \bigg]\nonumber\\
&=\E\bigg[ \frac{\partial \log \P_{\theta^*,S_j}\big[e_{j,a} | V(e_{j,a}) = \C', \{e_{j,a'}\}_{a' < a}\big]}{\partial \theta^*_i} \bigg|  V(e_{j,a})=\C', \{e_{j,a'}\}_{a' < a} \bigg]\nonumber\\
&= \sum_{\substack{e_{j,a} : V(e_{j,a}) = \C'\\ \{e_{j,a'}\}_{a' < a}}}\P_{\theta^*,S_j}\Big[e_{j,a} \big| V(e_{j,a}) =\C',\{e_{j,a'}\}_{a' < a}  \Big]  \; \frac{\partial}{\partial \theta^*_i} \log \P_{\theta^*,S_j}\Big[e_{j,a} \big| V(e_{j,a})=\C',\{e_{j,a'}\}_{a' < a} \Big] \nonumber\\
&=\frac{\partial}{\partial \theta^*_i} \sum_{e_{j,a} : V(e_{j,a}) = \C'}  \P_{\theta^*,S_j}\Big[e_{j,a} \big|  V(e_{j,a}) = \C' \Big]  =  \frac{\partial}{\partial \theta^*_i} 1 = 0 \nonumber\,,
\end{align}
where we used $\{e_{j,a} : V(e_{j,a}) = \C'\} = \{e_{j,a} : V(e_{j,a}) = \C',\{e_{j,a'}\}_{a' < a} \}$ which follows from the definition of rank-breaking edges $e_{j,a}$. This proves the desired claim.}

\subsection{Proof of Lemma \ref{lem:grad_bound}}

We view $\nabla \L_{\RB}(\theta^*)$ as the final value of a discrete time vector-valued martingale with values in $\reals^d$. Define $\nabla\L_{\RB}^{(e_{j,a})} \in \reals^d$ as the gradient vector arising out of each rank-breaking edge $\{e_{j,a}\}_{j \in [n], a\in [\ell_j]}$ as
\begin{eqnarray}
\nabla_i\L_{\RB}^{(e_{j,a})}(\theta^*) & \equiv & \sum_{\substack{\C \subseteq S_j}} \I\big\{ V(e_{j,a})= \{\C,i\} \big\}  \nabla_i \log \P_{\theta^*}(e_{j,a})\,, 
\end{eqnarray} 
such that $\nabla \L_{\RB}(\theta^*) = \sum_{j\in[n]} \sum_{a\in[\ell_j]} \nabla\L_{\RB}^{(e_{j,a})}$. 
We take $\nabla\L_{\RB}^{(e_{j,a})}$ as the incremental random vector in a martingale of $\sum_{j=1}^n \ell_j$ time steps. 
Let $H_{j,a}$ denote (the sigma algebra of) the history up to $e_{j,a}$ and define a sequence of random vectors in $\reals^d$: 
\begin{eqnarray*}
	Z_{j,a} &\equiv& \E[\nabla \L_{\RB}^{(e_{j,a})}(\theta^*) | H_{j,a} ]\;, 
\end{eqnarray*}
with the convention that $Z_{1,1}=\E[\nabla \L^{(e_{j,a})}_{\RB}(\theta^*)]=0$ as proved in Lemma \ref{lem:grad_expec}. 
It also follows from 
Lemma \ref{lem:grad_expec} that 
$\E[Z_{j,a+1}|Z_{j,a} ]= Z_{j,a}$ for $a<\ell_j$. 
Also, from the independence of samples, it follows that 
$\E[Z_{j+1,1}|Z_{j,\ell_j}] = Z_{j,\ell_j}$. 
Applying a generalized version of the vector Azuma-Hoeffding inequality which readily follows from [Theorem 1.8, \cite{Hay05}], we have
\begin{eqnarray}
\P\big[\,\norm{\nabla\L_{\RB}(\theta^*)} \geq \delta \,\big] & \leq & 
2e^3\exp\Bigg({-\frac{\delta^2}{\sum_{j=1}^n\sum_{a=1}^{\ell_j}m_{j,a} 2\gamma_2^{-1}e^{2b} }}\Bigg)\,, 
\end{eqnarray}  
where we used 
$\norm{\nabla\L_{\RB}^{(e_{j,a})} }^2 \leq m_{j,a} 2 \gamma_2^{-1} e^{2b}$. 
Choosing $\delta = \gamma_2^{-1} e^{b} \sqrt{6 np\log d}$ gives the desired bound. 

Now we are left to show that 
$\norm{\nabla\L_{\RB}^{(e_{j,a})}}^2 \leq 2m_{j,a} \gamma_2^{-1}  e^{2b} $ 
for any $\theta \in \Omega_b$. 
{Recall that $\sigma \in \Lambda_{T(e_{j,a})}$ is the set of all full rankings over $T(e_{j,a})$ items. 
In rest of the proof, with a slight abuse of notations, 
we extend each of these ranking $\sigma$  over $T(e_{j,a})\cup B(e_{j,a})$ items in the following way.   
Consider any full ranking $\tilde{\sigma}$ over $B(e_{j,a})$ items. Then for each $\sigma \in \Lambda_{T(e_{j,a})}$, the extension is such that
$\sigma(|T(e_{j,a})|+ c) = \tilde{\sigma}(c)$ for $ 1 \leq c \leq |B(e_{j,a})|$. The choice of ranking $\tilde{\sigma}$ will have no impact on any of the following mathematical expressions.}  
From the definition of $\P_{\theta}(e_{j,a})$ \eqref{eq:pair}, we have, for any $i \in V(e_{j,a})$,
\begin{align}\label{eq:grad_F}
\frac{\partial \P_{\theta}(e_{j,a})}{\partial \theta_i} \; =  &\; \I\{i \in T(e_{j,a})\}\P_{\theta}(e_{j,a})   \\ \nonumber 
& - \underbrace{\sum_{\sigma \in  \Lambda_{T(e_{j,a})}} \underbrace{\frac{\exp\left(\sum_{c = 1}^{m_{j,a}} \theta_{\sigma(c)} \right)}{\prod_{u=1}^{m_{j,a}} \left(\sum_{c'=u}^{r_{j,a}} \exp\left(\theta_{\sigma(c')}\right) \right)}}_{\equiv \, A_{\sigma}} \underbrace{\left( \sum_{u' = 1}^{m_{j,a}} \frac{\I{\{\sigma^{-1}(i) \geq u' \}}\exp(\theta_i)}{\sum_{c'=u'}^{r_{j,a}} \exp\left(\theta_{\sigma(c')}\right)} \right)}_{\equiv \,B_{\sigma,i}}}_{\equiv \,E_i}    \,.
\end{align}

Note that $A_{\sigma}, B_{\sigma,i}$ and $E_{i}$ depend on $e_{j,a}$. Observe that for any $1 \leq u' \leq m_{j,a}$ and any $\sigma \in \Lambda_{T(e_{j,a})}$,
\begin{eqnarray} \label{eq:grad_sum_zero}
\sum_{i \in V(e_{j,a})}{\I{\{\sigma^{-1}(i) \geq u' \}}\exp(\theta_i)} =  {\sum_{c'=u'}^{r_{j,a}} \exp\left(\theta_{\sigma(c')}\right)} \,.
\end{eqnarray}  
Therefore, $\sum_{i \in V(e_{j,a})} B_{\sigma,i} = m_{j,a}$. It follows that
\begin{eqnarray}\label{eq:grad_sum_1}
\sum_{i \in V(e_{j,a})} E_{i} \;=\; \sum_{\sigma \in \Lambda_{T(e_{j,a})}} A_{\sigma} \bigg( \sum_{i \in V(e_{j,a})} B_{\sigma,i}\bigg)\;\; =\;\; m_{j,a} \sum_{\sigma \in \Lambda_{T(e_{j,a})}} A_{\sigma} \;\;=\;\; m_{j,a} \P_{\theta}(e_{j,a})\,,
\end{eqnarray} 
where the last equality follows from the definition of $\P_{\theta}(e_{j,a})$ \eqref{eq:estimate}. Also, since for any $i,i'$, $e^{(\theta_i - \theta_{i'})} \leq e^{2b}$; for any $i$, 
$B_{\sigma,i} \leq e^{2b} \sum_{k = r_{j,a}-m_{j,a}+1}^{r_{j,a}} (1/k) \leq e^{2b}(1+\log(r_{j,a}/(r_{j,a}-m_{j,a}+1))) \leq \gamma_2^{-1} e^{2b}$, 
where the last inequality follows from the definition of $\gamma_2$ \eqref{eq:gamma12}
and the fact that $x\leq \sqrt{1+\log x}$ for all $x\geq1$. Therefore, 
$E_{i} \leq \gamma_2^{-1} e^{2b} \sum_{\sigma \in \Lambda_{T(e_{j,a})}} A_{\sigma} = \gamma_2^{-1} e^{2b}\P_{\theta}(e_{j,a})$. 
We have
$\partial \log \P_{\theta}(e_{j,a})/\partial \theta_i = 
(1/\P_{\theta}(e_{j,a})) \partial  \P_{\theta}(e_{j,a})/\partial \theta_i  = \I\{i \in T(e_{j,a})\} - E_{i} / \P_{\theta}(e_{j,a})$. 
Since $|T(e_{j,a})| = m_{j,a}$, 
$\norm{\nabla\L_{\RB}^{(e_{j,a}) }}^2 \leq m_{j,a} + \sum_{i\in V(e_{j,a})} (E_{i}/ \P_{\theta}(e_{j,a}))^2 \leq  2 m_{j,a}\gamma_2^{-1} e^{2b}$, 
where we used \eqref{eq:grad_sum_1} and the fact that $\gamma_2^{-1} \geq 1$.
  


\subsubsection{Proof of Lemma \ref{lem:hess_bound}}
First, we prove \eqref{eq:hess_def}. 
For brevity, remove $\{j,a\}$ from $\P_{\theta}(e_{j,a})$.
From Equations \eqref{eq:grad_F} and \eqref{eq:grad_sum_1}, and $|T(e_{j,a})| = m_{j,a}$, we have $\sum_{i \in V(e_{j,a})} \frac{\partial}{\partial \theta_i} \P_{\theta}(e)= m_{j,a}\P_{\theta}(e)-m_{j,a}\P_{\theta}(e) = 0$.  It follows that
\begin{align}\label{eq:defHessian}
&\sum_{i \in V(e_{j,a})} \bigg(\frac{\partial^2 \log \P_{\theta}(e)}{\partial\theta_{i'}\partial\theta_{i}}\bigg) = \nonumber\\
&\frac{1}{\P_{\theta}(e)}\frac{\partial }{\partial \theta_{i'}}\Bigg(\sum_{i \in V(e_{j,a})} \bigg(\frac{\partial\P_{\theta}(e)}{\partial \theta_i }\bigg)\Bigg) - \frac{1}{\left(\P_{\theta}(e)\right)^2}\frac{\partial \P_{\theta}(e)}{\partial \theta_{i'}} \Bigg(\sum_{i \in V(e_{j,a})} \bigg(\frac{\partial \P_{\theta}(e)}{\partial \theta_{i}}\bigg)\Bigg) = 0\,.
\end{align} 
Since by definition $\L_{\RB}(\theta) = \sum_{j = 1}^n \sum_{a =1}^{\ell_j} \log \P_{\theta}(e_{j,a})$, and $H_{ii'}(\theta) = \frac{\partial^2 \L_{\RB}(\theta)}{\partial\theta_i\partial\theta_{i'}}$ which is a symmetric matrix, Equation \eqref{eq:defHessian} implies that it can be expressed as given in Equation \eqref{eq:hess_def}. It follows that all-ones is an eigenvector of $H(-\theta)$ with the corresponding eigenvalue being zero. 

To get a lower bound on $\lambda_2(-H(\theta))$, we  apply Weyl's inequality 
\begin{eqnarray}
\lambda_2(-H(\theta)) \geq \lambda_2(\E[-H(\theta)]) - \norm{H(\theta) - \E[H(\theta)]}\,.
\end{eqnarray}
We will show in \eqref{eq:h5} that $\lambda_2(\E[-H(\theta)]) \geq e^{-6b}\alpha\gamma_1\gamma_2\gamma_3(np/(4(d-1)))$ and in \eqref{eq:h_error} that $\norm{H(\theta) - \E[H(\theta)]} \leq 16e^{4b}\nu \sqrt{\frac{p_{\max}}{\kappa_{\min}}\frac{np}{\beta(d-1)} \log d}$. Putting these together, 
\begin{eqnarray}
\lambda_2(-H(\theta)) &\geq & e^{-6b}\alpha\gamma_1\gamma_2\gamma_3\frac{np}{4(d-1)} - 16e^{4b}\nu \sqrt{\frac{p_{\max}}{\kappa_{\min}}\frac{np}{\beta(d-1)} \log d} \\
&\geq& \frac{e^{-6b}\alpha\gamma_1\gamma_2\gamma_3}{8}\frac{np}{(d-1)}\,,
\end{eqnarray}
where the last inequality follows from the assumption on $n\kappa_{\min}$ given in \eqref{eq:h_cond}.

To prove a lower bound on $\lambda_2(\E[-H(\theta)])$, we claim that for $\theta \in \Omega_{b}$,
\begin{eqnarray}\label{eq:h5}
\E\big[-H(\theta)\big] &\succeq & e^{-6b}\gamma_1\gamma_2\gamma_3 \sum_{j=1}^n  \frac{p_j}{4\kappa_j(\kappa_j-1)} \sum_{i < i' \in S_j}  (e_i - e_{i'})(e_i - e_{i'})^\top \\
&=& \frac{e^{-6b}\gamma_1\gamma_2\gamma_3}{4}L\,,\nonumber
\end{eqnarray}
where $L \in \cS^d$ is defined in \eqref{eq:L_comp}. Using $\lambda_2(L) = np\alpha/(d-1)$ from \eqref{eq:alphabeta}, we have $\lambda_2(-H(\theta)) \geq e^{-6b}\alpha\gamma_1\gamma_2\gamma_3(np/(4(d-1)))$. 
To prove \eqref{eq:h5}, notice that 
\begin{eqnarray} \label{eq:hess_offdiag_expec}
	\E[-H(\theta)_{ii'}] &=& \E\Big[ \sum_{j\in[n]} \sum_{a\in[\ell_j]} \I\big\{ (i,i')\subseteq V(e_{j,a}) \big\}\,  \frac{\partial^2 \log \P_{\theta}(e_{j,a})}{\partial\theta_i\partial\theta_{i'}}\Big]\;, 
\end{eqnarray}  
when $i\neq i'$. 
We will show that for any $i\neq i' \in V(e_{j,a})$,
\begin{eqnarray} \label{eq:h4}
\frac{\partial^2 \log \P_{\theta}(e_{j,a})}{\partial\theta_i\partial\theta_{i'}} &\geq&
\begin{cases}
 \frac{e^{-2b} m_{j,a}}{r_{j,a}^2} &\;\text{if}\;\; i,i' \in B(e_{j,a})\vspace{0.5em}\\
 -\frac{e^{4b}m_{j,a}^2}{(r_{j,a}-m_{j,a}+1)^2}  &\;\text{otherwise}\,.
\end{cases}
\end{eqnarray} 
We need to bound the probability of two items appearing in the bottom-set  $B(e_{j,a})$ and in the top-set $T(e_{j,a})$.
\begin{lemma} \label{lem:prob_bounds}
Consider a ranking $\sigma$ over a set $S \subseteq [d]$ such that $|S| = \kappa$. 
For any two items $i,\i \in S$, $\theta\in\Omega_b$, and  $1 \leq \ell,\ell_1,\ell_2 \leq \kappa-1$,
\begin{eqnarray} \label{eq:posl_lowerbound_eq}
\P_{\theta}\big[\sigma^{-1}(i), \sigma^{-1}(\i) > \ell \big] &\geq& \frac{e^{-4b}(\kappa-\ell)(\kappa-\ell-1)}{\kappa(\kappa-1)} \bigg(1 - \frac{\ell}{\kappa}\bigg)^{2e^{2b} -2} \;, \label{eq:prob1}\\
	\P_{\theta}\big[ \sigma^{-1}(i) = \ell \big] &\leq&  \frac{e^{6b}}{\kappa-\ell} \;, \label{eq:prob2}\\
	 	\P_{\theta}\big[ \sigma^{-1}(i) = \ell_1, \sigma^{-1}(i') = \ell_2 \big] &\leq&  \frac{e^{10b}}{(\kappa-\ell_1-1)(\kappa-\ell_2)}  \label{eq:prob3}\,.
\end{eqnarray}
where the probability $\prob_\theta$ is with respect to the sampled ranking resulting from PL weights $\theta\in\Omega_b$.
\end{lemma}
Substituting $\ell = \kappa_j-r_{j,a}+m_{j,a}$ in \eqref{eq:prob1}, and $\ell, \ell_1, \ell_2 \leq \kappa_j-r_{j,a}+m_{j,a}$ in \eqref{eq:prob2} and \eqref{eq:prob3}, we have,
\begin{eqnarray}
	\P_{\theta}\big[(i,i') \subseteq B(e_{j,a})\big] &\geq& \frac{e^{-4b}(r_{j,a}-m_{j,a})^2}{4\kappa_j(\kappa_j-1)} \, \Big(\frac{ r_{j,a}-m_{j,a}}{ \kappa_j} \Big)^{2e^{2b}-2} \label{eq:h1}\,,\\
	\P_{\theta}\big[i \in T(e_{j,a}), i' \in B(e_{j,a})\big] &\leq & m_{j,a}\max_{\ell \in [\kappa_j-r_{j,a}+m_{j,a}]}\prob(\sigma^{-1}(i)=\ell) \nonumber\\ 
	&\leq& \frac{e^{6b}m_{j,a}}{r_{j,a}-m_{j,a}}  \label{eq:h2}\,,\\
	\P_{\theta}\big[(i,i') \subseteq T(e_{j,a})\big] 
	&\leq& m_{j,a}^2 \max_{\ell_1,\ell_2\in[\kappa_j-r_{j,a}+m_{j,a}]} \prob(\sigma^{-1}(i)=\ell_1 , \sigma^{-1}(i') = \ell_2)\nonumber\\
	&\leq& \frac{e^{10b}m_{j,a}^2}{2\,(r_{j,a}-m_{j,a}-1)(r_{j,a}-m_{j,a})} \label{eq:h3}\,,
\end{eqnarray}
where \eqref{eq:h1} uses $r_{j,a} - m_{j,a}-1 \geq (r_{j,a} - m_{j,a})/4$, \eqref{eq:h2} uses $\P_{\theta}[i \in T(e_{j,a}), i' \in B(e_{j,a})] \leq \P_{\theta}[i \in T(e_{j,a})]$, and \eqref{eq:h2}-\eqref{eq:h3} uses counting on the possible choices. The bound in \eqref{eq:h3} is smaller than the one in \eqref{eq:h2} as per our assumption that $\gamma_3 > 0$.

Using Equations \eqref{eq:hess_offdiag_expec}-\eqref{eq:h4} and \eqref{eq:h1}-\eqref{eq:h3}, and the definitions of $\gamma_1,\gamma_2,\gamma_3$ from Section \ref{sec:topology}, we get 
\begin{align}
	&\E[-H(\theta)_{ii'}] \geq \nonumber\\
	&  \sum_{j\in[n]} \sum_{a\in[\ell_j]} \Big\{ \underbrace{\Big(\frac{r_{j,a}-m_{j,a}}{\kappa_j}\Big)^{2e^{2b}-2}}_{\geq\gamma_1} \underbrace{\Big(\frac{r_{j,a}-m_{j,a}}{r_{j,a}}\Big)^2}_{\geq \gamma_2} \frac{e^{-6b}m_{j,a}}{4 \kappa_j(\kappa_j-1)}\, 
	\;- \;\frac{e^{6b}m_{j,a}}{r_{j,a}-m_{j,a}}  \,\frac{e^{4b} m_{j,a}^2}{ (r_{j,a}-m_{j,a}+1)^2}\Big\} \nonumber\\
	&\geq  \sum_{j,a} \frac{\gamma_1\gamma_2e^{-6b}m_{j,a}}{4\kappa_j(\kappa_j-1)} \; \underbrace{\Big(1 \,-\, \frac{4e^{16b}}{\gamma_1}\frac{m_{j,a}^2 r_{j,a}^2 \kappa_j^2}{(r_{j,a}-m_{j,a})^5} \Big)}_{\geq \gamma_3}\;. 
\end{align}
This combined with \eqref{eq:hess_def} proves  the desired claim \eqref{eq:h5}. Further, in Appendix \ref{app:app1}, we show that if $m_{j,a} \leq 3$ for all $\{j,a\}$ then $\frac{\partial^2 \log \P_{\theta}(e_{j,a})}{\partial\theta_i\partial\theta_{i'}}$ is non-negative even for $i \neq i'\in T(e_{j,a})$, and $i \in T(e_{j,a}), i' \in B(e_{j,a})$ as opposed to a negative lower-bound given in \eqref{eq:h4}. Therefore, bound on $\E[-H(\theta)]$ in \eqref{eq:h5} can be tightened by a factor of $\gamma_3$.    

To prove claim \eqref{eq:h4}, define the following for $\sigma \in \Lambda_{T(e_{j,a})}$,
\begin{align} \label{eq:h_notations}
&A_{\sigma} \equiv \frac{\exp\left(\sum_{c = 1}^{m_{j,a}} \theta_{\sigma(c)} \right)}{\prod_{u=1}^{m_{j,a}} \left(\sum_{c'=u}^{r_{j,a}} \exp\left(\theta_{\sigma(c')}\right) \right)}\,,
B_{\sigma} \equiv \sum_{u' = 1}^{m_{j,a}} \frac{1}{\sum_{c'=u'}^{r_{j,a}} \exp\left(\theta_{\sigma(c')}\right)}\,, \nonumber\\
&B_{\sigma,i} \equiv \sum_{u' = 1}^{m_{j,a}} \frac{\I{\{\sigma^{-1}(i) \geq u' \}}}{\sum_{c'=u'}^{r_{j,a}} \exp\left(\theta_{\sigma(c')}\right)}\,,
C_{\sigma} \equiv \sum_{u' = 1}^{m_{j,a}} \frac{1}{\left(\sum_{c'=u'}^{r_{j,a}} \exp\left(\theta_{\sigma(c')}\right)\right)^2}\,,\nonumber\\
&C_{\sigma,i} \equiv \sum_{u' = 1}^{m_{j,a}} \frac{\I{\{\sigma^{-1}(i) \geq u' \}}}{\left(\sum_{c'=u'}^{r_{j,a}} \exp\left(\theta_{\sigma(c')}\right)\right)^2}\,,
C_{\sigma,i,i'} \equiv \sum_{u' = 1}^{m_{j,a}} \frac{\I{\{\sigma^{-1}(i),\sigma^{-1}(i') \geq u' \}}}{\left(\sum_{c'=u'}^{r_{j,a}} \exp\left(\theta_{\sigma(c')}\right)\right)^2}\,.
\end{align}
First, a few observations about the expression of $A_{\sigma}$. For any $\sigma \in \Lambda_{T(e_{j,a})}$ and any $i \in V(e_{j,a})$,  $\theta_i$ is in the numerator if and only if $i \in T(e_{j,a})$, since in all the rankings that are consistent with the observation $e_{j,a}$, $T(e_{j,a})$ items are ranked in top $m_{j,a}$ positions. For any $\sigma \in \Lambda_{T(e_{j,a})}$ and  any $i \in B(e_{j,a})$, $\theta_i$ is in all the product terms $\prod_{u=1}^{m_{j,a}}(\cdot)$ of the denominator, since in all the consistent rankings these items are ranked below $m_{j,a}$ position. For any $i \in T(e_{j,a})$, $\theta_i$ appears in product term corresponding to index $u$ if and only if item $i$ is ranked at position $u$ or lower than $u$ in the ranking $\sigma \in \Lambda_{T(e_{j,a})}$. 
Now, observe that $B_{\sigma}$ is defined such that the partial derivative of $A_{\sigma}$ with respect to any $i \in B(e_{j,a})$ is $-A_{\sigma}B_{\sigma}e^{\theta_i}$, and $B_{\sigma,i}$ is defined such that the partial derivative of $A_{\sigma}$ with respect to any $i \in T(e_{j,a})$ is $A_{\sigma}-A_{\sigma}B_{\sigma}e^{\theta_i}$. 
Further, observe that $-C_{\sigma}e^{\theta_i}$ is the partial derivative of $B_{\sigma}$ with respect to $i \in B(e_{j,a})$, $-C_{\sigma,i}e^{\theta_{i}}$ is the partial derivative of $B_{\sigma,i}$ with respect to $i \in T(e_{j,a})$, and $-C_{\sigma,i}e^{\theta_{i'}}$ is the partial derivative of $B_{\sigma,i}$ with respect to $i' \in B(e_{j,a})$. $-C_{\sigma,i,i'}e^{\theta_{i'}}$ is the partial derivative of $B_{\sigma,i}$ with respect to $i' \neq i \in T(e_{j,a})$. 

For ease of notation, we omit subscript $(j,a)$ whenever it is clear from the context. Also, we use $\sum_{\sigma}$ to denote $\sum_{\sigma \in \Lambda_{T(e_{j,a})}}$. With the above defined notations, from \eqref{eq:estimate}, we have, $\P_{\theta}(e) = \sum_{\sigma}A_{\sigma}$. 
With the above given observations for the notations in \eqref{eq:h_notations}, first partial derivative of $\P_{\theta}(e)$ can be expressed as following:
\begin{eqnarray} \label{eq:derivative}
\frac{\partial\P_{\theta}(e)}{\partial \theta_i} =
\begin{cases}
 \sum_{\sigma} \big(A_{\sigma} - A_{\sigma}B_{\sigma,i}e^{\theta_i} \big) &\quad\text{if}\;\; i \in T(e_{j,a})\\
\sum_{\sigma} \big( - A_{\sigma}B_{\sigma}e^{\theta_i}\big) &\quad\text{if}\;\; i \in B(e_{j,a})\,.
\end{cases}
\end{eqnarray}  
It follows that for $i \neq i' \in V(e_{j,a})$,
\begin{align} \label{eq:sec_derivative}
&\frac{\partial^2\P_{\theta}(e)}{\partial \theta_i \partial \theta_{i'}}  \nonumber\\
&=\begin{cases}
\sum_{\sigma}\big((A_{\sigma}(B_{\sigma})^2 + A_{\sigma}C_{\sigma})e^{(\theta_i+\theta_{\i})}\big) &\text{if}\; i,i' \in B(e_{j,a})\\
\sum_{\sigma} \big(A_{\sigma} - A_{\sigma}B_{\sigma,i'}e^{\theta_{i'}}  + (A_{\sigma}B_{\sigma,i}B_{\sigma,i'}+A_{\sigma}C_{\sigma,i,i'}) e^{(\theta_{i}+\theta_{i'})} - A_{\sigma}B_{\sigma,i}e^{\theta_i} \big) &\text{if}\; i,i'\in T(e_{j,a})\\
\sum_{\sigma} \big((A_{\sigma}B_{\sigma}B_{\sigma,i} +  A_{\sigma}C_{\sigma,i})e^{(\theta_i+\theta_{\i})}
 - A_{\sigma}B_{\sigma}e^{\theta_{\i}}\big) &\text{otherwise}\,.
\end{cases}
\end{align} 
Using $\frac{\partial^2 \log \P_{\theta}(e)}{\partial\theta_i\partial\theta_{i'}} = \frac{1}{\P_{\theta}(e)}\frac{\partial^2\P_{\theta}(e)}{\partial \theta_i \partial \theta_{i'}} - \frac{1}{\left(\P_{\theta}(e)\right)^2}\frac{\partial \P_{\theta}(e)}{\partial \theta_i}\frac{\partial \P_{\theta}(e)}{\partial \theta_{i'}}$, with above derived first and second derivatives, and after following some algebra, we have
\begin{align}\label{eq:h7}
&\frac{\left(\P_{\theta}(e)\right)^2}{e^{(\theta_i+\theta_{\i})}}\frac{\partial^2 \log \P_{\theta}(e)}{\partial\theta_i\partial\theta_{i'}} \nonumber\\
&=\begin{cases}
(\sum_{\sigma} A_{\sigma}) (\sum_{\sigma} A_{\sigma}(B_{\sigma})^2) - (\sum_{\sigma} A_{\sigma}B_{\sigma})^2 + (\sum_{\sigma}A_{\sigma})(\sum_{\sigma} A_{\sigma}C_{\sigma}) &\;\text{if}\;\; i,i' \in B(e_{j,a})\vspace{0.5em}\\
(\sum_{\sigma} A_{\sigma})(\sum_{\sigma} A_{\sigma}B_{\sigma,i}B_{\sigma,i'} + A_{\sigma}C_{\sigma,i,i'}) -  (\sum_{\sigma} A_{\sigma}B_{\sigma,i}) (\sum_{\sigma} A_{\sigma}B_{\sigma,i'}) &\;\text{if}\;\; i,i'\in T(e_{j,a})\vspace{0.5em}\\
(\sum_{\sigma} A_{\sigma})(\sum_{\sigma} A_{\sigma}B_{\sigma}B_{\sigma,i} + A_{\sigma}C_{\sigma,i}) -  (\sum_{\sigma} A_{\sigma}B_{\sigma}) (\sum_{\sigma} A_{\sigma}B_{\sigma,i}) &\;\text{otherwise}\,. 
\end{cases}
\end{align} 
Observe that from Cauchy-Schwartz inequality $(\sum_{\sigma} A_{\sigma}) (\sum_{\sigma} A_{\sigma}(B_{\sigma})^2) - (\sum_{\sigma} A_{\sigma}B_{\sigma})^2 \geq 0$. Also, we have $e^{(\theta_i+\theta_{\i})}C_{\sigma} \geq e^{-2b}(m/r^2)$ and $e^{\theta_i}B_{\sigma,i} \leq e^{\theta_i}B_{\sigma} \leq e^{2b}(m/(r-m+1))$ for any $i \in V(e_{j,a})$. This proves the desired claim \eqref{eq:h4}.

Next we need to upper bound deviation of $-H(\theta)$ from its expectation. From \eqref{eq:h7}, we have, $\big|\frac{\partial^2 \log \P_{\theta}(e_{j,a})}{\partial\theta_i\partial\theta_{i'}}\big| \leq 3e^{4b}m_{j,a}^2/(r_{j,a}-m_{j,a}+1)^2 \leq 3e^{4b}\nu m_{j,a}/(\kappa_j(\kappa_j-1))$, where the last inequality follows from the definition of $\nu$ \eqref{eq:nu12}. Therefore,  
\begin{eqnarray} \label{eq:h9}
-H(\theta) &\preceq & 3e^{4b} \nu \sum_{j=1}^n \sum_{a= 1}^{\ell_j}  \sum_{i < i' \in S_j} \I\{(i,i') \subseteq V(e_{j,a})\} \frac{m_{j,a}}{\kappa_j(\kappa_j-1)} (e_i - e_{i'})(e_i - e_{i'})^\top \\
&\preceq & 3e^{4b} \nu \sum_{j=1}^n  \sum_{i < i' \in S_j}    \frac{\sum_{a= 1}^{\ell_j} m_{j,a}}{\kappa_j(\kappa_j-1)} (e_i - e_{i'})(e_i - e_{i'})^\top\equiv \sum_{j=1}^n y_j L_j\,,
\end{eqnarray}
where $y_j = (3e^{4b}\nu p_j)/(\kappa_j(\kappa_j-1))$ 
and 
$L_j = \sum_{i<i'\in S_j}(e_i-e_{i'})(e_i-e_{i'})^\top = \kappa_j {\rm diag}(e_{S_j}) - e_{S_j}e_{S_j}^\top$ for $e_{S_j} = \sum_{i\in S_j}e_i$. Observe that $\norm{y_j L_j} \leq (3e^{4b}\nu p_{\max})/\kappa_{\min}$. Moreover, $L_j^2 \preceq \kappa_j L_j$, 
and it follows that 
\begin{eqnarray}
\sum_{j=1}^n y_j^2L_j^2 &\preceq& 9e^{8b} \nu^2 \sum_{j=1}^n \frac{ p_{j}^2 }{\kappa_j^2 (\kappa_j-1)^2} \kappa_j L_j   \preceq  \frac{9e^{8b}\nu^2p_{\max}}{\kappa_{\min}} L\;,
\end{eqnarray}
where we used the fact that $L = (p_j/(\kappa_j(\kappa_j-1)))\sum_{j=1}^n L_j$, for $L$ defined in \eqref{eq:L_comp}. 
Using $\lambda_d(L) = np/(\beta(d-1))$ from \eqref{eq:alphabeta}, it follows that $\norm{\sum_{j=1}^n \E_{\theta}[y_j^2Y_j^2]} \leq  \frac{9e^{8b}\nu^2p_{\max}}{\kappa_{\min}}\frac{np}{\beta(d-1)}$. By the matrix Bernstien inequality, with probability at least $1 - d^{-3}$,
\begin{eqnarray}\label{eq:h_error}
\norm{H(\theta) - \E[H(\theta)]} & \leq & 12e^{4b}\nu \sqrt{\frac{ p_{\max}}{\kappa_{\min}}\frac{np}{\beta(d-1)} \log d} + \frac{8e^{4b}\nu p_{\max} \log d}{\kappa_{\min}} \nonumber\\
 &\leq&  16e^{4b}\nu \sqrt{\frac{p_{\max}}{\kappa_{\min}}\frac{np}{\beta(d-1)} \log d}\,,
\end{eqnarray}  
where the last inequality follows from the assumption on $n\kappa_{\min}$ given in \eqref{eq:h_cond}.


\subsection{Proof of Lemma \ref{lem:prob_bounds}} 
{\textbf{Claim \eqref{eq:prob1}:}} Since providing a lower bound on $\P_{\theta}\big[\sigma^{-1}(i), \sigma^{-1}(\i) > \ell \big] $ 
for arbitrary $\theta$ is challenging, we construct a new set of parameters $\{\ltheta_j\}_{j\in[d]}$ from the original $\theta$. 
These new parameters are constructed such that  it is both easy to 
compute the probability and also provides a lower bound on the original distribution.
Define $\widetilde{\alpha}_{i,i',\ell,\theta}$ as
	\begin{align} \label{eq:posl_alpha}
	\widetilde{\alpha}_{i,i',\ell,\theta} \;\;  \equiv \;\; 
	\max_{\ell'\in[\ell]} \max_{\substack{\Omega \subseteq S\setminus\{i,\i\} \\ : |\Omega| = \kappa-\ell'}} \Bigg\{\frac{\exp(\theta_i)+\exp(\theta_{\i})}{\big(\sum_{j\in \Omega} 	\exp(\theta_j)\big)/|\Omega|} \Bigg \}\;,
	\end{align} 
and $\alpha_{i,i',\ell,\theta} = \ceil{\widetilde{\alpha}_{i,i',\ell,\theta}}$. For ease of notation we remove the subscript from $\alpha$ and $\widetilde{\alpha}$. We denote the sum of the weights by  $W \equiv \sum_{j \in S} \exp(\theta_j)$. 
We define a new set of parameters $\{\ltheta_j\}_{j \in S}$: 
\begin{eqnarray}
	\ltheta_j &=& \left\{ \begin{array}{rl}  
		\log(\lalpha/2) &\; \text{for} \; j = i \text{ or }\i\;, \\
		0&\;\text{otherwise}\;. \end{array}\right. 
\end{eqnarray}
Similarly define $\widetilde{W} \equiv \sum_{j \in S} \exp(\ltheta_j) = \kappa-2+\lalpha$.  We have,
\begin{eqnarray}\label{eq:posl_3}
&& \P_{\theta}\Big[\sigma^{-1}(i), \sigma^{-1}(\i) > \ell \Big] \nonumber\\
&=& \sum_{\substack{j_1 \in S \\ j_1 \neq i,\i}} \Bigg(\frac{\exp(\theta_{j_1})}{W}  \sum_{\substack{j_2 \in S \\ j_2 \neq i,\i,j_1}} \Bigg(\frac{\exp(\theta_{j_2})}{W-\exp(\theta_{j_1})}\cdots \Bigg( \sum_{\substack{j_{\ell} \in S \\ j_{\ell} \neq i,\i, \\ j_1,\cdots,j_{\ell-1}}} \frac{\exp(\theta_{j_{\ell}})}{W-\sum_{k=j_1}^{j_{\ell-1}}\exp(\theta_{k})} \Bigg) \cdots\Bigg)\Bigg) \nonumber\\
&=& \sum_{\substack{j_1 \in S \\ j_1 \neq i,\i}} \Bigg( \frac{\exp(\theta_{j_1})}{W-\exp(\theta_{j_1})} \cdots \sum_{\substack{j_{\ell-1} \in S \\ j_{\ell-1} \neq i,\i, \\ j_1,\cdots,j_{\ell-2}}} \Bigg( \frac{\exp(\theta_{j_{\ell-1}})}{W-\sum_{k=j_1}^{j_{\ell-1}}\exp(\theta_{k})}
\sum_{\substack{j_{\ell} \in S \\ j_{\ell} \neq i,\i, \\ j_1,\cdots,j_{\ell-1}}} \Bigg( \frac{\exp(\theta_{j_{\ell}})}{W}\Bigg)\cdots \Bigg)\Bigg) \nonumber\\ 
\end{eqnarray}
Consider the second-last summation term in the above equation and let
$\Omega_\ell = S\setminus\{i,i',j_1,\ldots,j_{\ell-2}\}$.
Observe that, $|\Omega_\ell| = \kappa-\ell$ and from equation \eqref{eq:posl_alpha}, $\frac{\exp(\theta_{i})+\exp(\theta_{\i})}{\sum_{j \in \Omega_\ell} \exp(\theta_j)} \leq \frac{\lalpha}{\kappa-\ell}$. We have, 
\begin{eqnarray}
&&\sum_{j_{\ell-1} \in \Omega_\ell} \frac{\exp(\theta_{j_{\ell-1}})}{W-\sum_{k=j_1}^{j_{\ell-1}}\exp(\theta_{k})} \nonumber\\
&=& \sum_{j_{\ell-1} \in \Omega_\ell } \frac{\exp(\theta_{j_{\ell-1}})}{W-\sum_{k=j_1}^{j_{\ell-2}}\exp(\theta_{k}) - \exp(\theta_{j_{\ell-1}})} \nonumber\\
&\geq& \frac{\sum_{j_{\ell-1} \in \Omega_\ell}\exp(\theta_{j_{\ell-1}})}{W-\sum_{k=j_1}^{j_{\ell-2}}\exp(\theta_{k})-\big(\sum_{j_{\ell-1} \in \Omega_\ell}\exp(\theta_{j_{\ell-1}})\big)/|\Omega_\ell|} \label{eq:posl_jensen_ineq}\\
&=& \frac{\sum_{j_{\ell-1} \in \Omega_\ell}\exp(\theta_{j_{\ell-1}})}{\exp(\theta_i)+\exp(\theta_{\i}) +\sum_{j_{\ell-1} \in \Omega_\ell}\exp(\theta_{j_{\ell-1}})-\big(\sum_{j_{\ell-1} \in \Omega_\ell}\exp(\theta_{j_{\ell-1}})\big)/|\Omega_\ell|} \nonumber\\
&=&\Bigg({\frac{\exp(\theta_{i})+\exp(\theta_{\i})}{\sum_{j_{\ell-1} \in \Omega_\ell} \exp(\theta_{j_{\ell-1}})} + 1 - \frac{1}{\kappa-\ell}}\Bigg)^{-1} \nonumber\\
&\geq& \Bigg(\frac{\lalpha}{\kappa-\ell} + 1 - \frac{1}{\kappa-\ell}\Bigg)^{-1} \label{eq:posl_1}\\
&=& \frac{\kappa-\ell}{\lalpha + \kappa-\ell-1} = \sum_{j_{\ell-1} \in \Omega_\ell } \frac{\exp(\ltheta_{j_{\ell-1}})}{\widetilde{W}-\sum_{k=j_1}^{j_{\ell-1}}\exp(\ltheta_{k})} \label{eq:posl_2}\;,
\end{eqnarray}
where \eqref{eq:posl_jensen_ineq} follows from the Jensen's inequality and the fact that for any $c >0$, $0 < x < c$, $\frac{x}{c-x}$ is convex in $x$. Equation \eqref{eq:posl_1} follows from the definition of $\widetilde{\alpha}_{i,i',\ell,\theta}$, \eqref{eq:posl_alpha}, and the fact that $|\Omega_\ell| = \kappa-\ell$. Equation \eqref{eq:posl_2} uses the definition of $\{\ltheta_j\}_{j \in S}$.

Consider $\{\Omega_{\widetilde{\ell}}\}_{2 \leq \widetilde{\ell} \leq \ell - 1}$, $|\Omega_{\widetilde{\ell}}| = \kappa - \widetilde{\ell}$, corresponding to the subsequent summation terms in \eqref{eq:posl_3}.  Observe that $\frac{\exp(\theta_i)+\exp(\theta_{\i})}{\sum_{j \in \Omega_{\widetilde{\ell}}} \exp(\theta_j)} \leq \alpha/|\Omega_{\widetilde{\ell}}|$. Therefore, each summation term  in equation  \eqref{eq:posl_3} can be lower bounded by the corresponding term where $\{\theta_j\}_{j \in S}$ is replaced by $\{\ltheta_j\}_{j \in S}$. Hence, we have 
\begin{eqnarray} \label{eq:posl_4}
&&\P_{\theta}\Big[\sigma^{-1}(i), \sigma^{-1}(\i) > \ell \Big] \nonumber\\
&\geq& \sum_{\substack{j_1 \in S \\ j_1 \neq i,\i}} \Bigg( \frac{\exp(\ltheta_{j_1})}{\widetilde{W}-\exp(\ltheta_{j_1})} \cdots \sum_{\substack{j_{\ell-1} \in S \\ j_{\ell-1} \neq i,\i, \\ j_1,\cdots,j_{\ell-2}}} \Bigg( \frac{\exp(\ltheta_{j_{\ell-1}})}{\widetilde{W}-\sum_{k=j_1}^{j_{\ell-1}}\exp(\ltheta_{k})} \sum_{\substack{j_{\ell} \in S \\ j_{\ell} \neq i,\i, \\ j_1,\cdots,j_{\ell-1}}} \Bigg( \frac{\exp(\theta_{j_{\ell}})}{W}\Bigg)\cdots \Bigg)\Bigg)  \nonumber\\
&\geq&  e^{-4b} \sum_{\substack{j_1 \in S \\ j_1 \neq i,\i}} \Bigg( \frac{\exp(\ltheta_{j_1})}{\widetilde{W}-\exp(\ltheta_{j_1})} \cdots \sum_{\substack{j_{\ell-1} \in S \\ j_{\ell-1} \neq i,\i, \\ j_1,\cdots,j_{\ell-2}}} \Bigg( \frac{\exp(\ltheta_{j_{\ell-1}})}{\widetilde{W}-\sum_{k=j_1}^{j_{\ell-1}}\exp(\ltheta_{k})} \sum_{\substack{j_{\ell} \in S \\ j_{\ell} \neq i,\i, \\ j_1,\cdots,j_{\ell-1}}} \Bigg( \frac{\exp(\ltheta_{j_{\ell}})}{\widetilde{W}}\Bigg)\cdots \Bigg)\Bigg)  \nonumber\\
&=& \big(e^{-4b}\big) \P_{\ltheta}\Big[\sigma^{-1}(i), \sigma^{-1}(\i) > \ell \Big] \;.
\end{eqnarray}
The second inequality uses $\frac{\exp(\theta_i)}{W} \geq e^{-2b}/\kappa$ and $\frac{\exp(\ltheta_i)}{\widetilde{W}} \leq e^{2b}/\kappa$. 
Observe that $\exp(\ltheta_j) = 1$ for all $j \neq i,\i$ and $\exp(\ltheta_i) + \exp(\ltheta_{\i}) = \widetilde{\alpha} \leq \ceil{\widetilde{\alpha}} = \alpha \geq 1$. Therefore, we have
\begin{eqnarray}
\P_{\ltheta}\Big[\sigma^{-1}(i), \sigma^{-1}(\i) > \ell \Big] 
&=& {\kappa-2 \choose \ell} \frac{\ell \,!}{(\kappa-2+\widetilde{\alpha})(\kappa-2+\widetilde{\alpha} - 1)\cdots(\kappa-2+\widetilde{\alpha} - (\ell-1))} \nonumber\\
&\geq& \frac{(\kappa-2)!}{(\kappa-\ell-2)!} \frac{1}{(\kappa + \alpha-2)(\kappa + \alpha - 3)\cdots(\kappa + \alpha - (\ell+1))} \nonumber\\
& \geq& \frac{(\kappa-\ell + \alpha-2)(\kappa -\ell +\alpha-3)\cdots (\kappa -\ell -1)}{(\kappa+\alpha-2)(\kappa+\alpha-3)\cdots(\kappa-1)} \nonumber\\
&\geq& \frac{(\kappa-\ell)(\kappa-\ell-1)}{\kappa(\kappa-1)}  \bigg( 1- \frac{\ell}{\kappa+1}\bigg)^{\alpha-2}\,. \label{eq:posl_6}
\end{eqnarray}
Claim \eqref{eq:prob1} follows by combining Equations \eqref{eq:posl_4} and \eqref{eq:posl_6} and using the fact that $\alpha \leq 2e^{2b}$.\\
{\textbf{Claim \eqref{eq:prob2}:}}
Define,  
\begin{align} \label{eq:posll_upperbound_eq}
\widetilde{\alpha}_{\ell,\theta} \;\; \equiv \;\; \min_{i \in S} \min_{\ell' \in [\ell]} \min_{\substack{\Omega \in S\setminus\{i\} \\ : |\Omega| = \kappa-\ell'+1}} \Bigg\{\frac{\exp(\theta_i)}{\big(\sum_{j\in \Omega} \exp(\theta_j)\big)/|\Omega|} \Bigg \}\;.
\end{align} 
Also, define $\alpha_{\ell,\theta} \equiv \floor{\widetilde{\alpha}_{\ell,\theta}}$. Note that $\alpha_{\ell,\theta} \geq 0$ and $\widetilde{\alpha}_{\ell,\theta} \leq e^{2b}$. We denote the sum of the weights by  $W \equiv \sum_{j \in S} \exp(\theta_j)$. 
Analogous to the proof of claim \eqref{eq:prob1}, we define the new set of parameters $\{\ltheta_j\}_{j \in S}$: 
\begin{eqnarray}
	\ltheta_j &=& \left\{ \begin{array}{rl}  
		\log(\widetilde{\alpha}_{\ell,\theta}) &\; \text{for} \; j = i \;, \\
		0&\;\text{otherwise}\;. \end{array}\right. 
\end{eqnarray}
Similarly define $\widetilde{W} \equiv \sum_{j \in S} \exp(\ltheta_j) = \kappa-1+\widetilde{\alpha}_{\ell,\theta}$. Using the techniques similar to the ones used in proof of claim \eqref{eq:prob1}, we have, 
\begin{eqnarray}
\P_{\theta}\Big[\sigma^{-1}(i) = \ell \Big] & \leq & e^{4b} \P_{\ltheta}\Big[\sigma^{-1}(i) = \ell\Big]\,.
\label{eq:posl_upper2}
\end{eqnarray}
Observe that $\exp(\ltheta_j) = 1$ for all $j \neq i$ and $\exp(\ltheta_i) = \widetilde{\alpha}_{\ell,\theta} \geq \floor{\widetilde{\alpha}_{\ell,\theta}} = \alpha_{\ell,\theta} \geq 0$. Therefore, we have
\begin{eqnarray}
\P_{\ltheta}\Big[\sigma^{-1}(i) = \ell \Big] 
&=& {\kappa-1 \choose \ell-1} \frac{\lalpha_{\ell,\theta}(\ell-1)!}{(\kappa-1+\widetilde{\alpha}_{\ell,\theta})(\kappa-2+\widetilde{\alpha}_{\ell,\theta})\cdots(\kappa-\ell+\widetilde{\alpha}_{\ell,\theta}) } \nonumber\\
&\leq& \frac{(\kappa-1)!}{(\kappa-\ell)!} \frac{e^{2b}}{(\kappa -1 + \alpha_{\ell,\theta})(\kappa -2+ \alpha_{\ell,\theta} )\cdots(\kappa -\ell + \alpha_{\ell,\theta} )} \nonumber\\
&\leq& \frac{e^{2b}}{\kappa}  \bigg( 1- \frac{\ell}{\kappa+\alpha_{\ell,\theta}}\bigg)^{\alpha_{\ell,\theta}-1} \leq \frac{e^{2b}}{\kappa-\ell}\,. \label{eq:posl_upper6}
\end{eqnarray}
Claim \ref{eq:prob2} follows by combining Equations \eqref{eq:posl_upper2} and \eqref{eq:posl_upper6}.\\
{\textbf{Claim \eqref{eq:prob3}:}}
Again, we construct a new set of parameters $\{\ltheta_j\}_{j\in[d]}$ from the original $\theta$ using $\lalpha_{\ell,\theta}$ defined in \eqref{eq:posll_upperbound_eq}:
\begin{eqnarray}
	\ltheta_j &=& \left\{ \begin{array}{rl}  
		\log(\widetilde{\alpha}_{\ell,\theta}) &\; \text{for} \; j \in \{i,i'\} \;, \\
		0&\;\text{otherwise}\;. \end{array}\right. 
\end{eqnarray}
Similarly define $\widetilde{W} \equiv \sum_{j \in S} \exp(\ltheta_j) = \kappa-2 + 2\widetilde{\alpha}_{\ell,\theta}$.  Using the techniques similar to the ones used in proof of claim \eqref{eq:prob1}, we have, 
\begin{eqnarray}\label{eq:posll_upper2}
\P_{\theta}\Big[\sigma^{-1}(i) = \ell_1, \sigma^{-1}(i') = \ell_2 \Big]  & \leq & e^{8b} \P_{\ltheta}\Big[\sigma^{-1}(i) = \ell_1, \sigma^{-1}(i') = \ell_2 \Big] 
\end{eqnarray}
Observe that $\exp(\ltheta_j) = 1$ for all $j \neq i,\i$ and $\exp(\ltheta_i) = \exp(\ltheta_{i'}) = \widetilde{\alpha}_{\ell,\theta} \geq \floor{\widetilde{\alpha}}_{\ell,\theta} = \alpha_{\ell,\theta} \geq 0$. Therefore, we have
\begin{align}
&=\P_{\ltheta}\Big[\sigma^{-1}(i) = \ell_1, \sigma^{-1}(i') = \ell_2 \Big] \nonumber\\
&= \Bigg(\frac{{\kappa-2 \choose \ell_2-2} \lalpha_{\ell,\theta}^2(\ell_2-2)!}{(\kappa-2+2\lalpha_{\ell,\theta})(\kappa-1+2\lalpha_{\ell,\theta})\cdots(\kappa-2 +2\lalpha_{\ell,\theta} -(\ell_1-1))} \nonumber\\
&\frac{1}{(\kappa-2 +\lalpha_{\ell,\theta} -(\ell_1-1))\cdots(\kappa-2+\lalpha_{\ell,\theta} -(\ell_2-2)) }\Bigg) \nonumber\\
&\leq \frac{(\kappa-2)!}{(\kappa-\ell_2)!} \frac{e^{4b}}{(\kappa -2)(\kappa -1 )\cdots(\kappa -\ell_1-1)(\kappa -\ell_1-1)\cdots(\kappa-\ell_2)} \nonumber\\
&\leq \frac{e^{4b}}{(\kappa-\ell_1-1)(\kappa-\ell_2)}\,. \label{eq:posll_upper6}
\end{align}
Claim \ref{eq:prob3} follows by combining Equations \eqref{eq:posll_upper2} and \eqref{eq:posll_upper6}.

\subsection{Proof of Theorem \ref{thm:lower_bound}}
Let $H(\theta) \in \cS^d$ be Hessian matrix such that $H_{ii'}(\theta) = \frac{\partial^2\L_{\RB}(\theta)}{\partial \theta_i \partial \theta_{i'}}$. The Fisher information matrix is defined as $I(\theta) = - \E_{\theta}[H(\theta)]$. From lemma \ref{lem:concave}, $\L_{\RB}(\theta)$ is concave. This implies that $I(\theta)$ is positive-semidefinite and from \eqref{eq:hess_def} its smallest eigenvalue is zero with all-ones being the corresponding eigenvector. 
Fix any unbiased estimator $\widehat{\theta}$ of $\theta \in \Omega_b$. Since, $\widehat{\theta} \in \U$, $\widehat{\theta} - \theta$ is orthogonal to $\vect{1}$. The Cramer-Rao lower bound then implies that $\E[\norm{\widehat{\theta}-\theta^*}^2] \geq \sum_{i=2}^d \frac{1}{\lambda_i(I(\theta))}$. Taking supremum over both sides gives
\begin{eqnarray}
\sup_{\theta} \E[\norm{\widehat{\theta}-\theta^*}^2] & \geq & \sup_{\theta} \sum_{i=2}^d \frac{1}{\lambda_i(I(\theta))} \geq \sum_{i=2}^d \frac{1}{\lambda_{i}(I(\vect{0}))}\,.
\end{eqnarray}
 In the following, we will show that 
\begin{eqnarray} \label{eq:lb1}
	I(\vect{0}) \;=\; -\E_{\theta}[H(\vect{0})] 
	& \preceq & \sum_{j=1}^n \sum_{a=1}^{\ell_j} \frac{m_{j,a}-\eta_{j,a}}{\kappa_j(\kappa_j-1)} \sum_{i < i' \in S_j} (e_i-e_{i'})(e_i-e_{i'})^\top \\ 
	& \preceq & \max_{j,a} \big\{ m_{j,a}-\eta_{j,a}\big\} \, L  \,. \label{eq:lb3}
\end{eqnarray}
Using Jensen's inequality, we have $\sum_{i=2}^d \frac{1}{\lambda_{i}(I(\vect{0}))} \geq \frac{(d-1)^2}{\sum_{i=2}^d\lambda_{i}(I(\vect{0}))} = \frac{(d-1)^2}{\Tr(I(\vect{0}))}$. 
From \eqref{eq:lb1}, we have $\Tr(I(\vect{0})) \leq \sum_{j,a} (m_{j,a}-\eta_{j,a}) $.  
From \eqref{eq:lb3}, we have 
$\sum_{i=2}^d {1}/{\lambda_{i}(I(\vect{0}))} \geq (1/\max\{m_{j,a}-\eta_{j,a}\})\sum_{i=1}^d {1}/{\lambda_{i}(L)}$ .
This proves the desired claim. 

Now we are left to show claim \eqref{eq:lb1}. Consider a rank-breaking edge $e_{j,a}$.
Using notations defined in lemma \ref{lem:hess_bound}, in particular Equation \eqref{eq:h_notations}, and omitting subscript $\{j,a\}$ whenever it is clear from the context, we have, for any $i \in V(e_{j,a})$,
\begin{eqnarray}
\frac{\partial^2\P_{\theta}(e_{j,a})}{\partial^2 \theta_i} =
\begin{cases}
\sum_{\sigma}\big(-A_{\sigma}B_{\sigma}e^{\theta_i} + A_{\sigma}(B_{\sigma})^2e^{2\theta_i} + A_{\sigma}C_{\sigma}e^{\theta_i}\big) &\;\text{if}\;\; i \in B(e_{j,a})\\
\sum_{\sigma} \big( A_{\sigma}  - 3A_{\sigma}B_{\sigma,i}e^{\theta_i} +A_{\sigma}C_{\sigma,i}) e^{2\theta_{i}} + A_{\sigma}(B_{\sigma,i})^2e^{2\theta_i}\big) &\;\text{if}\;\; i\in T(e_{j,a})\,,
\end{cases}
\end{eqnarray}
and using \eqref{eq:derivative}, we have 
\begin{eqnarray}
\frac{\partial^2\log\P_{\theta}(e_{j,a})}{\partial^2 \theta_i}\Big|_{\theta = \vect{0}} =
\begin{cases}
\big((C_{\sigma} - B_{\sigma})\big)_{\theta = \vect{0}} &\;\text{if}\;\; i \in B(e_{j,a})\\
\big( \frac{1}{m_{j,a}!} \sum_{\sigma}\big(C_{\sigma,i} - B_{\sigma,i} + (B_{\sigma,i})^2 \big) - \big(\sum_{\sigma}\frac{B_{\sigma,i}}{m_{j,a}!}\big)^2\big)_{\theta = \vect{0}} &\;\text{if}\;\; i\in T(e_{j,a})\,,
\end{cases}
\end{eqnarray}
where $\sigma \in \Lambda_{T(e_{j,a})}$ and the subscript $\theta=0$ indicates the the respective quantities are evaluated at $\theta=0$. 
From the definitions given in \eqref{eq:h_notations}, for $\theta = \vect{0}$, we have $ B_{\sigma} - C_{\sigma} = \sum_{u=0}^{m-1}\frac{(r-u-1)}{(r-u)^2}$ and, $ \sum_{\sigma} (B_{\sigma,i} - C_{\sigma,i})/(m!) = \frac{1}{m}\sum_{u=0}^{m-1}\frac{(m-u)(r-u-1)}{(r-u)^2}$. Also, $\sum_{\sigma}B_{\sigma,i}/(m!) = \frac{1}{m}\sum_{u=0}^{m-1}\frac{m-u}{r-u}$ and $\sum_{\sigma} (B_{\sigma,i})^2/(m!) = \frac{1}{m}\sum_{u=0}^{m-1}\big(\sum_{u'=0}^u \frac{1}{r-u'}\big)^2$. Combining all these and, using 
$\P_{\theta = \vect{0}}[i \in T(e_{j,a})] = m/\kappa$ and $\P_{\theta = \vect{0}}[i \in B(e_{j,a})] = (r-m)/\kappa$, and after following some algebra, we have for any $i \in S_j$,
\begin{eqnarray} 
&&-\E\bigg[\frac{\partial^2\log\P_{\theta}(e_{j,a})}{\partial^2 \theta_i}\Big|_{\theta = \vect{0}} \bigg]   \nonumber\\
& =&\frac{1}{\kappa}\bigg(m - \sum_{u=0}^{m-1} \frac{1}{r-u} -\frac{1}{m} \sum_{u=0}^{m-1} \frac{u(m-u)}{(r-u)^2} - \frac{1}{m} \sum_{u=0}^{m-2} \frac{2u}{r-u}\bigg(\sum_{u'>u}^{m-1}\frac{m-u'}{r-u'}\bigg)\Bigg) \nonumber\\ 
&=& \frac{m_{j,a}-\eta_{j,a}}{\kappa_{j}} \,, \label{eq:lb2}
\end{eqnarray}
where $\eta_{j,a}$ is defined in \eqref{eq:eta}. 
Since row-sums of $H(\theta)$ are zeroes, \eqref{eq:hess_def}, and for $\theta = \vect{0}$, all the items are exchangeable, we have for any $i \neq i' \in S_j$, 
\begin{eqnarray}
\E\bigg[\frac{\partial^2 \log \P_{\theta}(e_{j,a})}{\partial\theta_i\partial \theta_{i'}} \Big|_{\theta = \vect{0}}\bigg] &  = & \frac{m_{j,a}-\eta_{j,a}}{\kappa_j(\kappa_j-1)}\,,
\end{eqnarray}
The claim \eqref{eq:lb1} follows from the expression of $H(\theta)$, Equation \eqref{eq:hess_def}.

To verify \eqref{eq:lb2}, 
observe that $(r-m)(B_{\sigma} - C_{\sigma}) + m (\sum_{\sigma}B_{\sigma,i}/(m!)) = m - \sum_{u=0}^{m-1} \frac{1}{r-u}$. And,
\begin{eqnarray*}
&&\frac{1}{m} \bigg(\sum_{u=0}^{m-1}\frac{m-u}{r-u} \bigg)^2 - \sum_{u=0}^{m-1}\bigg(\sum_{u'=0}^{u} \frac{1}{r-u'}\bigg)^2\\
 &=&\sum_{u=0}^{m-1}\bigg(\frac{(m-u)^2}{m(r-u)^2} - \frac{m-u}{(r-u)^2} \bigg) + \sum_{0 \leq u < u' \leq m-1} \bigg(\frac{2(m-u)(m-u')}{m(r-u)(r-u')} - \frac{2(m-u')}{(r-u)(r-u')} \bigg)\\
 &=&\sum_{u=0}^{m-1} \frac{-u(m-u)}{m(r-u)^2} + \sum_{0 \leq u < u' \leq m-1} \frac{-2u(m-u')}{m(r-u)(r-u')}\,.
\end{eqnarray*}


\subsection{Tightening of Lemma \ref{lem:hess_bound}} \label{app:app1}
Recall that $\P_{\theta}(e_{j,a})$ is same as probability of $\P_{\theta}[T(e_{j,a}) \succ B(e_{j,a})]$ that is the probability that an agent ranks $T(e_{j,a})$ items above $B(e_{j,a})$ items when provided with a set comprising $V(e_{j,a})$ items. As earlier, for brevity of notations, we omit subscript $\{j,a\}$ whenever it is clear from the context. For $m = 1$ or $2$, it is easy to check that all off-diagonal elements in hessian matrix of $\log\P_{\theta}(e)$ are non-negative. However, since number of terms in summation in $\P_{\theta}(e)$ grows as $m!$, for $m \geq 3$ the straight-forward approach becomes too complex. Below,  we derive expressions for cross-derivatives in hessian, for general $m$, using alternate definition (sorting of independent exponential r.v.'s in increasing order) of PL model, where the number of terms grow only as $2^m$. However, we are unable to analytically prove that the cross-derivatives are non-negative for $m >2$. Feeding these expressions in MATLAB and using symbolic computation, for $m=3$, we can simplify these expressions and it turns out that they are sum of only positive numbers. For $m=4$, with limited computational power it becomes intractable. We believe that it should hold for any value of $m < r$.
Using \eqref{eq:h4}, we need to check only for cross-derivatives for the case when $i \neq i' \in T(e_{j,a})$ or $i \in T(e_{j,a}), i' \in B(e_{j,a})$. Since, minimum of exponential random variables is exponential, we can assume that $|B(e_{j,a})| =1$ that is $r = m+1$. Define $\lambda_i \equiv e^{\theta_i}$. Without loss of generality, assume $T(e_{j,a}) = \{2,\cdots,m+1\}$ and $B(e_{j,a}) = \{1\}$. Define $C_x = \prod_{i=3}^{m+1} (1-e^{-\lambda_i x})$. Then, using the alternate definition of the PL model, we have, $\P_{\theta}(e) = \int_{0}^{\infty} C_x(1-e^{-\lambda_2 x})\lambda_1 e^{-\lambda_1 x}dx$. Following some algebra, $\frac{\partial^2 \log \P_{\theta}(e)}{\partial\theta_1 \partial \theta_2} \geq 0$ is equivalent to $A_1 \geq 0$, where $A_1 \equiv $ 
\begin{eqnarray*}
\bigg(\int C_x \big( xe^{-\lambda_1 x} - xe^{-\lambda x}\big)dx\bigg) \bigg(\int C_x xe^{-\lambda x}dx\bigg) - \bigg(\int C_x(e^{\lambda_1 x} - e^{-\lambda x})dx \bigg) \bigg(\int C_x x^2 e^{-\lambda x}dx\bigg)\,,
\end{eqnarray*}
where all integrals are from $0$ to $\infty$ and, $\lambda \equiv \lambda_1+\lambda_2$. Consider $A_1$ as a function of $\lambda_1$. Since $A_1(\lambda_1) = 0$ for $\lambda_1 = \lambda$, showing $\partial A_1/\partial \lambda_1 \leq 0$ for $0 \leq \lambda_1 \leq \lambda$ would suffice. Following some algebra, and using $\lambda_1 \leq \lambda$, $\partial A_1/\partial \lambda_1 \leq 0$ is equivalent to $A_2(\lambda_1) \equiv \big(\int_{0}^{\infty} C_x xe^{-\lambda_1 x}\big) / \big(\int_{0}^{\infty}C_x x^2e^{-\lambda_1 x}\big)$ being monotonically non-decreasing in $\lambda_1$. To further simplify the condition, define $f^{(0)}(y) = 1/y^2$, $g^{(0)}(y) = 1/y^3$ and, $f^{(1)}(y) = f^{(0)}(y) - f^{(0)}(y+\lambda_3)$, and recursively $f^{(m-1)}(y) = f^{(m-2)}(y) - f^{(m-2)}(y+\lambda_{m+1})$. Similarly define $g^{(0)}, \cdots, g^{(m-1)}$. Using these recursively defined functions,
\begin{eqnarray*}
2A_2(\lambda_1) &=&  \frac{f^{(m-1)}(\lambda_1)}{ g^{(m-1)}(\lambda_1)}\,,\;\; \nonumber\\
\text{for $m=3$},\;\; 2A_2(\lambda_1) &=& \frac{\lambda_1^{-2} - (\lambda_1+\lambda_3)^{-2} - (\lambda_1+\lambda_4)^{-2} + (\lambda_1+\lambda_3+\lambda_4)^{-2}}{\lambda_1^{-3} - (\lambda_1+\lambda_3)^{-3} - (\lambda_1+\lambda_4)^{-3} + (\lambda_1+\lambda_3+\lambda_4)^{-3}}\,.
\end{eqnarray*}     
Therefore, we need to show that $A_2(\lambda_1)$ is monotonically non-decreasing in $\lambda_1 \geq 0$ for any non-negative $\lambda_3, \cdots, \lambda_m$, and that would suffice to prove that the cross-derivatives arising from $i \in T(e_{j,a}), i' \in B(e_{j,a})$ are non-negative.
      
For cross-derivatives arising from $i \neq i' \in T(e_{j,a})$, define $B_x = \prod_{i=4}^{m+1}(1-e^{\lambda_i x})e^{-\lambda_1 x}$. $\frac{\partial^2 \log \P_{\theta}(e)}{\partial \theta_2 \partial \theta_3} \geq 0$ is equivalent to $A_3 \geq 0$, where $A_3 \equiv$
\begin{eqnarray*}
\bigg(\int B_x(1-e^{-\lambda_2 x})(1-e^{-\lambda_3 x})dx\bigg) \bigg(\int B_x x^2e^{-(\lambda_2 + \lambda_3)x} dx\bigg) \nonumber\\
- \bigg( \int B_x(1-e^{-\lambda_2 x}) xe^{-\lambda_3 x} dx\bigg)\bigg( \int B_x(1-e^{-\lambda_3 x}) xe^{-\lambda_2 x} dx\bigg)\,, 
\end{eqnarray*} 
where all integrals are from $0$ to $\infty$. For $m=3$, using MATLAB we can show that both types of cross-derivatives are non-negative. 

%
%
%
%


\section*{Acknowledgements} 
This work is supported by NSF SaTC award CNS-1527754, and NSF CISE award CCF-1553452.


 \bibliography{./_ranking,./_probability,./_machinelearning}

\end{document}